\pdfoutput=1
\documentclass[sigconf, nonacm]{acmart}

\newcommand{\spara}[1]{\vspace{.2cm}\noindent \textbf{#1}}

\newcommand{\hui}[1]{{\color{blue} #1 - Hui}}

\usepackage{microtype}
\usepackage{graphicx}
\usepackage{multirow}
\usepackage{tabularx}
\usepackage{array}
\usepackage{booktabs} 
\usepackage{caption}
\usepackage{subcaption}
\usepackage{amsmath}
\usepackage{bbding}
\usepackage{xcolor, pifont}
\usepackage{balance}
\usepackage{flushend}
\usepackage{float}
\usepackage[export]{adjustbox}
\newcommand*\colourxmark[1]{%
  \expandafter\newcommand\csname #1xmark\endcsname{\textcolor{#1}{\ding{55}}}%
}
\colourxmark{blue}
\colourxmark{green}
\colourxmark{red}
\colourxmark{black}
\newcommand{\revision}[1]{{\color{black}#1}}

\newcommand{\finalrevision}[1]{{#1}}

\newcommand*\colourcheck[1]{%
  \expandafter\newcommand\csname #1check\endcsname{\textcolor{#1}{\ding{51}}}%
}
\colourcheck{blue}
\colourcheck{green}
\colourcheck{red}
\colourcheck{black}

\begin{document}
\title{Graph Neural Network Training Systems: A Performance Comparison of Full-Graph and Mini-Batch}
\author{Saurabh Bajaj, Hojae Son, Juelin Liu, Hui Guan, and Marco Serafini}
\affiliation{
  \institution{University of Massachusetts, Amherst}
  \city{Amherst}
  \state{Massachusetts}
  \country{USA}
}
\email{{sbajaj, hojaeson, juelinliu, huiguan, mserafini}@umass.edu}


\begin{abstract}
Graph Neural Networks (GNNs) have gained significant
attention in recent years due to their ability to learn representations of graph-structured data. Two common methods for training GNNs are mini-batch training and full-graph training. 
Since these two methods require different training pipelines and systems optimizations, two separate classes of GNN training systems emerged, each tailored for one method. 
Works that introduce systems belonging to a particular category predominantly compare them with other systems within the same category, offering limited or no comparison with systems from the other category.
Some prior work also justifies its focus on one specific training method by arguing that it achieves higher accuracy than the alternative.
The literature, however, has incomplete and contradictory evidence in this regard.

In this paper, we provide a comprehensive empirical comparison of representative full-graph and mini-batch GNN training systems. 
We find that the mini-batch training systems consistently converge faster than the full-graph training ones across multiple datasets, GNN models, and system configurations.
We also find that mini-batch training techniques converge to similar to or often higher accuracy values than full-graph training ones, showing that mini-batch sampling is not necessarily detrimental to accuracy.
Our work highlights the importance of comparing systems across different classes, using time-to-accuracy rather than epoch time for performance comparison, and selecting appropriate hyperparameters for each training method separately.




\end{abstract}

\maketitle

\section{Introduction}
Graph neural networks (GNNs) are a class of machine learning models that reached state-of-the-art performance in many tasks related to the analysis of graph-structured data, including social network analysis, recommendations, and fraud detection \cite{zitnik_2018_modeling, schlichtkrull_modeling, hamilton_embedding}.
They are often used to process large graphs that have millions of vertices and billions of edges~\cite{pinsage,hu2021ogblsc,liu2023bgl}. 
A large volume of recent work, both in academia and industry, has been dedicated to scaling GNN training to support such large graphs using multi-GPU systems.
This is a challenging problem because GNNs run multiple rounds of \emph{message passing} across neighboring vertices, which is an irregular computation.

\spara{Two classes of GNN systems: Full-graph and mini-batch.}
GNNs can be trained using either a \emph{mini-batch} or a full-batch (typically called \emph{full-graph}) approach, much like other machine learning models.
\revision{
In standard deep neural network (DNN) training, the dataset consists of individual training examples that can be processed independently and have no structural dependencies.
In GNNs, in contrast, the training data is composed of vertices that are interconnected through edges, forming a graph structure where vertices cannot be treated as independent training examples. 
Full-graph and mini-batch training deal with these dependencies with different data management pipelines to partition data and parallelize computation and communication when scaling to large graphs.
}
This resulted in the development of two distinct classes of GNN training systems, each designed to support either mini-batch or full-graph training. 


\begin{figure}
    \centering
    \includegraphics[width=0.9 \columnwidth]{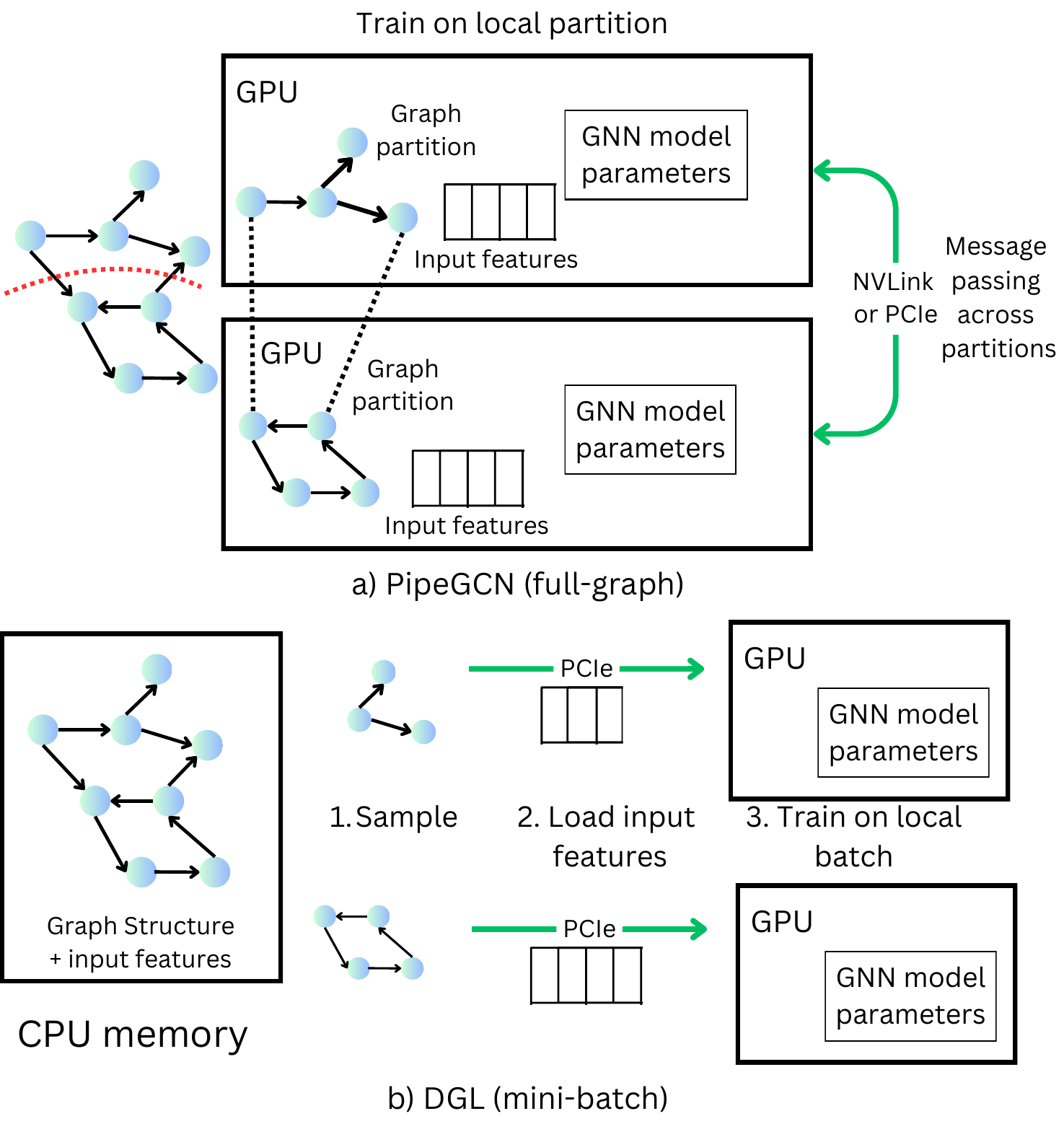}
    \caption{Different data management pipelines in two example systems: PipeGCN (full-graph) and DGL (mini-batch). The diagrams omit gradient synchronization.}
    \label{fig:fg_mb_training_diagram}
\end{figure}

Full-graph GNN training performs message-passing across the entire graph at each epoch.
To scale to large graphs that do not fit in the memory of a single GPU, multi-GPU full-graph training systems use \emph{model parallelism}: they partition the graph, process different partitions at different GPUs, and exchange hidden vertex features across partitions~\cite{DBLP_tripathy_cagnet, ma_neugraph, jia_improving_roc}.
For example, the PipeGCN~\cite{wan_pipegcn} full-graph training system partitions the input graph and keeps each partition in a different GPU (see Figure~\ref{fig:fg_mb_training_diagram}(a)).
The black dotted lines show the communication of hidden vertex features across GPUs  at each GNN layer in the forward and backward pass. 
The communication happens through either a fast cross-GPU bus such as NVLink, if available, or PCIe.

Mini-batch training breaks up the training set into mini-batches.
Each training epoch now consists of multiple iterations, one per mini-batch.
Mini-batch training is popular in production systems because it is amenable to scaling through \emph{data parallelism} \cite{zheng2021distdgl, zheng_2022_distributed,zhu2019aligraph}.
A unique challenge when applying data parallelism to GNNs is that preparing mini-batches requires expensive pre-processing at each iteration.
At each iteration, data-parallel mini-batch training systems sample
\revision{a subset of the k-hop neighbors}
of the vertices in the mini-batch, load each sample into a different GPU, and perform local message passing only within each sample, without passing message across GPUs.
For example, Figure ~\ref{fig:fg_mb_training_diagram}(b) shows a training iteration in the popular DGL system. 
First, each GPU obtains a sample for a subset of the vertices in the mini-batch.
Graph sampling is preferably executed using the GPU since it is a computationally expensive step in itself~\cite{nextdoor}.
Next, the sample and the input features of its vertices are loaded to the GPUs.
Finally, each GPU performs a forward and backward pass on the local sample independently. 

These different data management pipelines result in a different set of optimizations.
Full-graph training systems must deal with expensive and irregular computation and high communication costs at each epoch.
In mini-batch training systems, instead, computation at each iteration is much more lightweight and there is no cross-GPU exchange of hidden features at each layer.
However, pre-processing is a major bottleneck.
Samples are typically large, so sampling them and loading them into GPUs is expensive.

\revision{
The common rationale for using full-graph GNN training is to ensure that the training algorithm sees all the required dependent data in the graph, which is assumed to be the key to achieving high accuracy~\cite{thorpe_dorylus, wan_pipegcn, wan2022bnsgcn, G3_full_graph, DBLP_tripathy_cagnet, cai_2021_dgcl, yang_2023_betty, wan2023adaptive, sancus, mostafa2022sequential, ma_neugraph}.
Mini-batch GNN training introduces additional variability in the gradient estimation, even compared to regular mini-batch DNN training.
Specifically, in regular DNN training, the stochasticity in gradient estimation primarily arises from using mini-batches, which are random subsets of the training data, in each iteration. 
In mini-batch GNN training, the graph sampling used to form micro-batches is an additional source of stochasticity beyond the sampling of vertices in the mini-batch, because the training algorithm is not exposed to all the required dependent data in the graph.
}

\spara{Motivation: What is the state of the art?} 
\revision{
Despite their different approaches and bottlenecks, full-graph and mini-batch training systems share the same goal: efficient and accurate training of Graph Neural Networks (GNNs). 
It is therefore important to establish which GNN training systems, be they  full-graph or mini-batch systems, converge faster and to a higher accuracy under different scenarios.
There is, however, no clear answer to this question in the literature.
}
We performed an exhaustive review of the existing literature and found that 
prior works that propose systems in one class compare them with other systems in the same class, with limited or no performance evaluation of systems in the other class.
We report a detailed discussion of the evidence in the literature in Section ~\ref{sec:background}.
\emph{Overall, the existing literature makes it difficult to get a clear picture of the state-of-the-art of the field.}

\spara{Performance evaluation.}
In this paper, we address this open issue by performing a thorough performance comparison of representative state-of-the-art GNN training systems across the two classes.
For full-graph training, we consider PipeGCN~\cite{wan_pipegcn}, BNS-GCN~\cite{wan2022bnsgcn}, and AdaQP~\cite{adaqp-repo}.
For mini-batch training, we consider DGL~\cite{wang2019dgl}, DistDGL~\cite{zheng2021distdgl}, and Quiver~\cite{tan_quiver}, and three sampling algorithms: Neighborhood Sampling~\cite{hamilton2017inductive}, ClusterGCN~\cite{chiang_2019_clustergcn}, and GraphSaint~\cite{graphsaint-ipdps19}.

Our results show that \emph{the mini-batch training systems are consistently faster than the full-graph training counterparts in reaching the same target accuracy across all datasets, GNN models, and hardware deployments we considered}. 
These results are consistent with the evidence in the literature.
Several papers that propose full-graph training systems use \textit{epoch time}, \revision{which is the average time needed to execute one epoch}, as the main metric to compare with mini-batch training systems, 
and show that mini-batch systems have a larger epoch time ~\cite{wang_2022_neutronstar, G3_full_graph, md2021distgnn, chen2023gnnpipe}.
Our evaluation confirms these results.
However, our results also highlight that \textit{time-to-accuracy} is a better metric for comparison. 
Mini-batch training typically requires fewer epochs to converge because it updates the model parameters multiple times per epoch, once at each iteration, whereas full-graph training performs only one update per epoch.
Therefore, even if mini-batch training systems perform more work per epoch than full-graph training ones to run multiple iterations, their overall time-to-accuracy is lower.
Our empirical results show that this holds consistently across all systems, deployments, models, and datasets we consider. 
We report our performance results in Section~\ref{sec:perf_eval} \revision{and support and generalize our empirical observations with an analysis in Section~\ref{sec:analysis}.}

\spara{Performance vs. accuracy.}
Using a system with a longer time-to-accuracy can be justifiable if a model can converge to a \emph{higher} accuracy.
This is an important factor when navigating the performance trade-offs of using different GNN training systems.

Our results show that, with proper hyperparameter tuning, \emph{mini-batch training can reach a similar accuracy as, if not higher accuracy than, full-graph training across different datasets and models.}
In our evaluation, we matched or exceeded all the test accuracy results for the GNN models we considered in the literature we reviewed.
We observe that given a dataset and GNN model, hyperparameters that yield high accuracy for one training method do not work as well for the other.
This may explain the contradicting results found in the literature and underline the importance of conducting separate hyperparameter tuning for the two training methods.
\revision{ These empirical findings also imply that filtering information during graph aggregation is not necessarily detrimental to accuracy, which is against the assumption made by works advocating full-graph training approaches. 
}
We discuss these results in Section~\ref{sec:results_section}.

\spara{Lessons learned.}
Our results show that the mini-batch training systems we consider consistently achieve better performance than the full-graph training ones and have similar accuracy.
We stress, however, that the goal of this study is not to establish the inherent superiority of one class of systems over the other.
Our study is an empirical evaluation and the field is continuously evolving.
\revision{Mini-batch training also has limitations, such as a slightly higher variance in accuracy between training runs and a high variance across different sampling algorithms.}

\revision{Overall, our work highlights that existing algorithmic optimizations such as sampling or asynchrony can improve efficiency with no or minimal accuracy degradation, representing a promising research direction to achieve further performance and accuracy improvements.}
From a methodological perspective, our work highlights the importance of comparing GNN training systems across different classes and it also indicates a set of principles on how to perform this comparison.
\finalrevision{The source code, data, and other artifacts used for this evaluation are publicly available~\cite{mini-full-repo2}}.

\section{Background and Motivation} \label{sec:background}
This section provides the necessary background on GNN and GNN training systems to follow the rest of the paper. 
It then discusses the existing evidence in the literature.

\subsection{Graph Neural Networks} 
\label{sec:gnn-background}
Given a graph $G(V, E)$, where each node $v \in V $ is represented with a feature vector $h_{v}^0$ (input feature of vertex $v$) and the edges between any two vertices $u$ and $v$ is $e_{uv} \in E$. A GNN at $l^{th}$ layer performs the following computation to get the hidden features of a vector:
\begin{equation}
\label{eqn:GNN-layer}
    h_{v}^l = \Phi^l(h_{v}^{l-1}, \Psi^l(h_{u}^{l-1}, \forall u \in N \cup {v}), W^l), \\
\end{equation}
where $h_{v}^l$ are the $l^{th}$ layer features for vertex $v$, $N$ represents the incoming neighbor vertices of vertex $v$, $W^l$ is the model weight matrix, $\Psi^l$ is any aggregation function and $\Phi^l$ is an update function.

\newcolumntype{C}[1]{>{\centering\let\newline\\\arraybackslash\hspace{0pt}}m{#1}}
\newcolumntype{L}[1]{>{\raggedright\let\newline\\\arraybackslash\hspace{0pt}}m{#1}}
\newcolumntype{R}[1]{>{\raggedleft\let\newline\\\arraybackslash\hspace{0pt}}m{#1}}

\begin{table*}[]
    \caption{Review of literature reporting claims of how mini-batch and full-graph training compare. (Notation: TTA - time to accuracy, ACC - accuracy, FG - Full graph, MB - mini-batch, ET - Epoch time)}
    \label{tab:literature_review}
    \begin{center}
    \begin{footnotesize}
    \begin{tabular}{|l|l|L{3.5cm}|C{1.4cm}|c|l|}
    \hline
    Proposes & System / Paper &  Main FG vs. MB Claims & Experimental Evidence & Year & Evaluation Setup\\ 
    \hline
    \multirow{16}{*}{Full-graph} 
    &  GNNPipe \cite{chen2023gnnpipe}& FG lower ET & ET & 2023 & Multi-host multi-GPU\\
    \cline{2-6}
    &  Betty \cite{yang_2023_betty} & FG higher ACC & ACC & 2023 & Single-host single-GPU\\
    \cline{2-6}
    & HongTu \cite{wang2023hongtu} & MB information loss, MB lower ACC & ACC, TTA & 2023 & Single-host multi-GPU, distributed CPU-only \\
    \cline{2-6}
    & ADGNN \cite{song2023adgnn} & FG higher ACC, FG better convergence & \redxmark & 2023 & Multi-host CPU, Multi-host single-GPU \\
    \cline{2-6}
    &  G3 \cite{G3_full_graph}& FG higher ACC, lower ET & ACC, ET & 2023 & Multi-host multi-GPU\\
    \cline{2-6}
    &  AdaQP \cite{wan2023adaptive} & MB information loss & \redxmark & 2023 & Multi-host multi-GPU \\
    \cline{2-6}
    & PipeGCN \cite{wan_pipegcn}& FG higher ACC & ACC & 2022  & Multi-host multi-GPU\\
    \cline{2-6}   
    & BNS-GCN \cite{wan2022bnsgcn} & FG higher ACC, lower ET & ACC, ET & 2022 & Multi-host multi-GPU\\
    \cline{2-6}
    &  NeutronStar \cite{wang_2022_neutronstar}& FG lower ET & ET & 2022 & Multi-host single-GPU \\
    \cline{2-6}
    &  Sancus \cite{sancus}& MB information loss & \redxmark & 2022 & Multi-host multi-GPU\\
    \cline{2-6}
    &  SAR \cite{mostafa2022sequential} & MB noisy gradients & \redxmark & 2022 & Multi-host CPU \\
    \cline{2-6}
    & Dorylus \cite{thorpe_dorylus} & FG higher ACC & ACC & 2021 & Multi-host CPU, Multi-host single-GPU\\
    \cline{2-6}
    &  DistGNN \cite{md2021distgnn} & FG lower ET & ET & 2021 & Multi-host CPU\\
    \cline{2-6}
    &  DGCL \cite{cai_2021_dgcl} &  MB ACC loss & \redxmark & 2021 &  Multi-host multi-GPU \\
    \cline{2-6}
    &  CAGNET \cite{DBLP_tripathy_cagnet} & FG higher ACC & \redxmark & 2020 & Multi-host multi-GPU\\
    \cline{2-6}
    & ROC \cite{jia_improving_roc} & FG lower TTA & TTA & 2020 & Multi-host multi-GPU\\
    \cline{2-6}
    &  NeuGraph \cite{ma_neugraph} &  FG convergence guarantee & \redxmark & 2019 & Single-host multi-GPU \\

    \cline{1-6}
    \cline{1-6}

    \multirow{12}{*}{Mini-batch} 

    & BGL \cite{liu2023bgl} & MB and FB same performance & \redxmark & 2023 & Multi-host CPU, Multi-host multi-GPU\\
    \cline{2-6}
    &NeutronOrch \cite{ai2023neutronorch} & FG impractical, limited GPU memory & \redxmark & 2023 & Single-host multi-GPU\\
    \cline{2-6}
    &DistDGLv2 \cite{zheng_2022_distributed} & MB higher ACC, lower TTA & ACC, TTA & 2022 & Multi-host CPU, Multi-host multi-GPU\\
    \cline{2-6}
    &SALIENT \cite{kaler2022accelerating_salient}& MB lower ET, higher ACC & ACC, ET & 2022 & Multi-host multi-GPU\\
    \cline{2-6}
    &GNNLab \cite{yang_gnnlab} & FG hard to scale & \redxmark & 2022 & Single-host multi-GPU\\
    \cline{2-6}
    &ByteGNN \cite{byteGCN_zheng}& FG not practical for large graphs & \redxmark & 2022 & Multi-host CPU \\
    \cline{2-6}
    &CM-GCN \cite{cm_gcn_zhao}& MB lower TTA & ACC, TTA & 2021  & Multi-host CPU \\
    \cline{2-6}
    &Blocking based \cite{yao2021blocking} & FG higher memory and computational complexity & \redxmark & 2021 & - \\
    \cline{2-6}
    &DistDGL \cite{zheng2021distdgl} & MB lower TTA & \redxmark & 2020 & Multi-host CPU, multi-host multi-GPU\\
    \cline{2-6}
    &PaGraph \cite{lin_2020_pagraph} &  MB and FB same performance & \redxmark & 2020 & Single-host multi-GPU\\
    \cline{2-6}
    &Cluster-GCN \cite{chiang_2019_clustergcn} & FG - memory: bad; time per epoch: good; convergence: bad & \redxmark & 2019 & - \\
    \cline{2-6}
    & LADIES \cite{zou_2019_layerdependent} & MB higher ACC & ACC & 2019 & - \\
    \cline{1-6}
    \cline{1-6}

    \multirow{4}{*}{Other papers} 
    &Rethinking \cite{li2024rethinking}& FG not scalable & ACC & 2024 &  - \\
    \cline{2-6}
    &RDM \cite{rdm} & FG ACC higher or equal, MB faster convergence & ACC, TTA & 2023 &  -\\
    \cline{2-6}
    & EXACT \cite{liu_exact} & FG - higher ACC for small graphs , MB - higher ACC for large graphs  & ACC & 2021 &  - \\
    \cline{2-6}
    &OGB \cite{hu2020ogb} & MB higher ACC & ACC & 2020 & - \\

    \hline
    \end{tabular}
    \end{footnotesize}
    \end{center}
    \vskip -0.1in
\end{table*}

\subsection{Full-Graph vs. Mini-Batch Training: Different Systems for Different Pipelines} 
\label{sec:mb-vs-fg-optimizations}
Full-graph and mini-batch training systems implement different data management pipelines for scalable multi-GPU training, which resulted in the emergence of two separate categories of systems. 
\revision{In this paper, we focus our evaluation on GPU-based training on multi-GPU systems, which is a common configuration for scalable GNN training.}
\finalrevision{We consider system-level optimizations of the vanilla full-graph and mini-batch training pipelines that improve performance and support any GNN model, as defined in Section~\ref{sec:gnn-background}. We show the classes of optimizations we consider and the related baselines in Figure~\ref{fig:baselines}.}
For a comprehensive discussion on distributed GNN training systems, see also~\cite{shao2023distributed, lin2023comprehensive}.


\revision{
\spara{Full-graph training systems.}
The key idea of full-graph training is to execute the GNN layers for all vertices in the graph, as shown in Eqn.~\ref{eqn:GNN-layer}.
To scale to large graphs, distributed full-graph training systems partition the graph into multiple subgraphs such that each can fit into the memory of one GPU.
They use a \emph{model-parallel} approach to train across GPUs that requires exchanging vertex features across partitions. 
The main focus of work on full-graph training systems has been on reducing the cost of vertex feature communication.
We now discuss existing work on full-graph GNN training systems in terms of two design dimensions: \emph{training data management} and \emph{algorithmic optimizations} (see Figure~\ref{fig:baselines}).

Some full-graph training systems focused on optimizing the management of training data without introducing approximations that can potentially impact accuracy.
If there are not enough GPUs to store each partition in a different GPU, it is necessary to transfer data between the host memory and GPU memory.
Prior works such as NeuGraph~\cite{ma_neugraph}, RoC~\cite{jia_improving_roc}, G3~\cite{G3_full_graph}, and HongTu~\cite{wang2023hongtu} have proposed techniques to optimize host-GPU communication.
When there are enough GPUs to distribute all the partitions of the graph in GPU memory, these optimizations are not required and training proceeds as depicted in Figure~\ref{fig:fg_mb_training_diagram}(a)~\cite{wang_2022_neutronstar,G3_full_graph,cai_2021_dgcl,DBLP_tripathy_cagnet}.

Some recent works improve performance by further introducing algorithmic optimizations that can impact accuracy, unlike the work we described previously.
Sancus~\cite{sancus}, PipeGCN~\cite{wan_pipegcn}, and GNNPipe~\cite{chen2023gnnpipe} introduce \emph{asynchrony} and allow GPUs to operate on stale vertex features. 
This enables GPUs to overlap communication with computation instead of having to wait at each layer for fresh vertex features coming from other GPUs.
An alternative research direction has been to use \emph{sampling}.
BNS-GCN samples boundary nodes with edges across partitions at each epoch and exchanges vertex features only for those vertices~\cite{wan2022bnsgcn}.
ADGNN proposes an aggregation-difference sampling algorithm~\cite{song2023adgnn}.
Finally, the AdaQP system proposed \emph{message quantization} to reduce the communication cost, together with new partitioning algorithms~\cite{wan2023adaptive}.

\begin{figure}[t]
    \centering
    \includegraphics[width=\columnwidth]{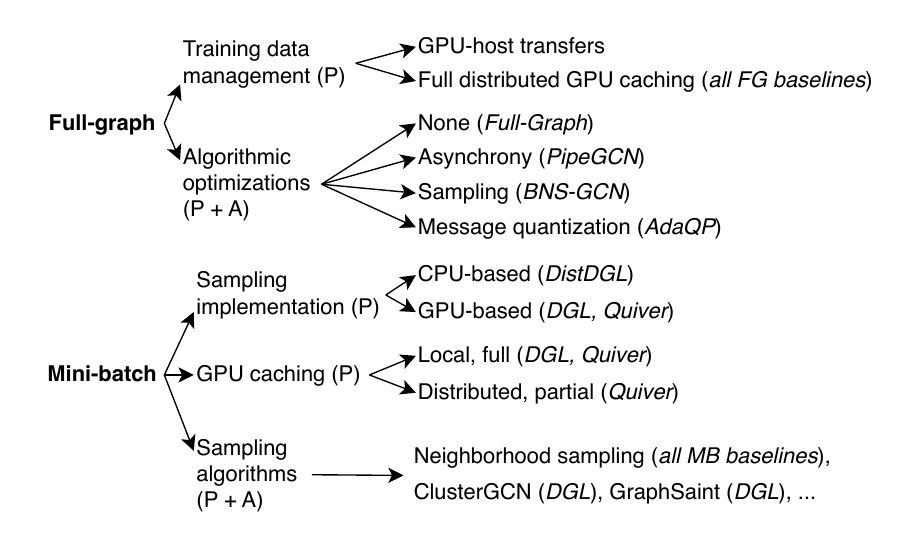}
    \caption{\revision{Classes of optimizations that impact performance (P) and accuracy (A), and representative systems evaluated in this work.}}
    \label{fig:baselines}
\end{figure}

}

\revision{
\spara{Mini-batch training systems.}
Mini-batch training systems break up the training dataset into mini-batches and train and update the model on one mini-batch at a time.
They use \emph{data parallelism} to scale to multiple GPUs.
Each GPU computes the hidden features of a subset of the vertices in the mini-batch, which is called the micro-batch.
In principle, to execute $k$ GNN layers without loss of information as shown in Eqn.~\ref{eqn:GNN-layer}, each GPU would need to load all the vertices in the $k$-hop neighborhood of the vertices in the micro-batch. 
The resulting subgraph, however, is often too large to be loaded into one GPU, inducing the so-called neighborhood explosion problem.
Mini-batch training systems use sampling to load only a subset of the $k$-hop neighborhood.
This sample-load-train pipeline is depicted in Figure~\ref{fig:fg_mb_training_diagram}(b).
Compared to full-graph training, data parallelism eliminates the need to exchange hidden vertex features across partitions.
The pipeline, however, introduces two different bottlenecks: sampling and data loading.
We now discuss systems-level optimizations that mitigate these bottlenecks without impacting accuracy and then discuss the choice of sampling algorithms (see Fig~\ref{fig:baselines}).

Initial mini-batch systems ran \emph{CPU-based sampling}, but this can represent a major performance bottleneck, motivating the need for \emph{GPU-based sampling}~\cite{nextdoor,case-sampling}.
Several systems-level optimizations have been proposed to run sampling algorithms on GPUs efficiently~\cite{nextdoor,c-saw,wang2021skywalker,gsampler,dsp_cai}.

In terms of management of the training data, loading the training data to the GPUs can be a significant performance bottleneck.
To address it, prior work proposed \emph{caching training data in GPU memory}, such as the input features~\cite{lin_2020_pagraph} or the graph structure~\cite{yang_gnnlab, ducati, dsp_cai}.
In \emph{local GPU caching}, each GPU only accesses its local cache~\cite{lin_2020_pagraph, yang_gnnlab}.
In \emph{distributed GPU caching} GPUs can directly access data cached in the main memory of other GPUs, which is faster than accessing data from the host memory when GPUs are connected through a fast bus like NVLink~\cite{tan_quiver,Yang2022WholeGraphAF,ugache,dsp_cai}.
Mini-batch training systems can leverage GPU caching also when only a subset of the training dataset is cached (\emph{partial caching}).
They load data from the host memory in case of cache misses~\cite{min2022graph,yang_gnnlab,legion_sun,tan_quiver,liu2023bgl,dsp_cai}.

In terms of algorithmic optimizations that can impact accuracy, mini-batch training systems can run different \emph{sampling algorithms}.
The seminal work that proposed mini-batch GNN training, GraphSage, proposed a simple neighborhood sampling approach~\cite{hamilton2017inductive}.
Subsequent work proposed many different algorithms such as FastGCN~\cite{chen2018fastgcn}, ClusterGCN~\cite{chiang_2019_clustergcn}, LADIES~\cite{zou_2019_layerdependent}, or GraphSaint~\cite{graphsaint-ipdps19}.
}

\subsection{Motivation} \label{sec:review}
The motivation for this work is that prior work on GNN training systems makes it difficult to establish the state of the art in the field.
A significant amount of work on GNN training do not compare or evaluate their systems against systems using a different training approach and training pipeline (full-graph or mini-batch)~\cite{kaler2023communicationefficient_salient++, legion_sun, dsp_cai, chen2018fastgcn, hamilton2017inductive, chen2018stochastic_vrgcn, huang2018adaptive_sgcn, graphsaint-ipdps19, pmlr-v198-gasteiger22a_influence_based, wang2021skywalker, zhu2019aligraph, zhang2020agl, kipf_semisupervised,p3_gandhi}.
\finalrevision{Some prior work focused on the mini-batch training pipeline of Figure~\ref{fig:fg_mb_training_diagram}(b), evaluating the impact of using different data partitioning algorithms, configurations of the neighborhood sampling algorithm, and data loading optimizations within the same system~\cite{comprehensive-gnn}.}
In the following, we review 33 papers on GNN training that make claims comparing systems in the two classes.
We summarize the claims in these papers in Table \ref{tab:literature_review}.

We classify the papers into three broad groups.
The first group consists of papers that propose full-graph training.
Out of the 17 papers in this group, 
10 provide experimental evidence. 
The rest 
7 papers make claims that full-graph training is superior in terms of higher accuracy, lower epoch time, and convergence guarantees and support the claims by only citing other papers. 
The second group includes 12 papers that propose mini-batch training. 
Among these papers, only 4 provide experimental evidence to support claims that full-graph training is difficult to scale and requires more memory, while mini-batch training has lower time-to-accuracy.
The last group consists of 4 papers that do not propose one specific training approach.
%
Overall, we find the following shortcomings in the existing literature, which motivate our work.

\spara{Conflicting claims about how the two approaches compare.}
Several papers proposing full-graph systems \cite{thorpe_dorylus, wan_pipegcn, wan2022bnsgcn, G3_full_graph, DBLP_tripathy_cagnet, cai_2021_dgcl, yang_2023_betty, wang2023hongtu, song2023adgnn} demonstrate that their systems achieve higher accuracy and lower epoch time compared to mini-batch systems. 
Some papers mention that mini-batch causes information loss, noisy gradients, and does not guarantee convergence \cite{wan2023adaptive, sancus, mostafa2022sequential, ma_neugraph, wang2023hongtu, song2023adgnn}
However, papers proposing mini-batch systems \cite{kaler2022accelerating_salient, zou_2019_layerdependent, zheng_2022_distributed, cm_gcn_zhao, zheng2021distdgl} and some other works \cite{hu2020ogb} present contradicting evidence and claims showing that mini-batch training leads to higher accuracy and faster convergence to the target accuracy level. Other works claim full-graph training is impractical, requires high memory, and is hard to scale \cite{ai2023neutronorch, yang_gnnlab, byteGCN_zheng, chiang_2019_clustergcn, yao2021blocking}.
Some argue that no approach is clearly superior and assessing their effectiveness requires a comprehensive evaluation which is out of the scope of their work~\cite{liu_exact, rdm, song_2021_network,liu2023bgl, lin_2020_pagraph, wang2023hongtu}.

\spara{Comparing performance only in terms of epoch time.}
Certain works compare full-graph and mini-batch training systems based only on epoch time~\cite{wang_2022_neutronstar, G3_full_graph, md2021distgnn, chen2023gnnpipe}. 
As we will show, a comparison in terms of time-to-accuracy is more informative given the differences in the two training approaches.
Our observations indicate that although mini-batch training exhibits higher epoch time, it still takes less time to reach a target accuracy level (Section \ref{sec:time_to_acc}).


\spara{Experiment design.}
Some works \cite{jia_improving_roc, liu_exact, li2024rethinking, kaler2022accelerating_salient} do not always use the same GNN model (e.g., GraphSAGE or GAT) for the two approaches as a basis of comparison. This conflates the effect of the training approach with the GNN model choice, since different models may exhibit varying performance even when trained with the same technique. Some papers do not mention explicitly whether they perform separate hyperparameter tuning for each method~\cite{G3_full_graph, wan_pipegcn, wan2022bnsgcn, thorpe_dorylus, cm_gcn_zhao, kaler2022accelerating_salient}.
As we show in our experimental evaluation of Section \ref{sec:results_section}, given a dataset and GNN model, it is possible to find hyperparameter settings where one method achieves higher accuracy than another and vice versa.
Using hyperparameter tuning is necessary to perform a fair comparison.




\begingroup
\renewcommand{\arraystretch}{0.8}

\begin{table}[]
\caption{Dataset statistics. (*) The input features and the training data for Orkut are synthetic.}
\label{tab:dataset_statistics}
    \begin{center}
    \resizebox{\columnwidth}{!}{%
    \begin{tabular}{|l*{5}{|c}|r|}

    \toprule
    Datasets & Nodes & Edges & Classes & Features & Train/Val/Test \\ 

    \midrule
    Pubmed &  19k & 108k &  3 & 500 & 18k/0.5k/1k \\
    Arxiv & 169k &  1.1M & 40 & 128 & 91k/29k/48k \\
    Reddit & 232k & 11.6M & 41 & 602 & 152k/23k/55k \\
    Products & 2.4M & 61.9M & 47 & 100 & 196K/49K/2.2M \\
    Orkut & 3M & 117M & 2* & 128* & 1.8M/614k/614k* \\ 
    Papers100M & 111M & 1.6B & 172 & 128 & 1.2M/125k/214k \\
    \bottomrule
    \end{tabular}%

    }
    \end{center}
\end{table}

\endgroup

\section{Experimental methodology} \label{sec:experimental_setup}

\textbf{Hardware setup.} We run our experiments using two types of hosts. 
The first is a lower-end one, which we call \emph{PCIe}. 
It has 2 Intel Xeon E5-2620 v3 CPUs with 12 cores each, 256GB of host memory, and 4 NVIDIA Tesla m40 GPUs, each having 24GB of memory, connected via PCIe.
These servers are connected with a 1 Gbps network.
The second host type is higher end and we call it \emph{NVLink}.
It has 2 Intel Xeon Platinum 8480+ CPUs with 56 cores each, 256 GB of host memory, and 4 A100 GPUs, each with 80GB of memory.
These servers are connected with a 25 Gbps network.
By default, all our experiments use 4 GPUs per host.

\spara{Datasets and models.} \label{sec:dataset}
Our experiments use six datasets:  pubmed \cite{namata:mlg12}, ogbn-arxiv \cite{hu2020ogb}, reddit \cite{hamilton2017inductive}, ogbn-products \cite{hu2020ogb}, orkut \cite{yang2012defining_snap} and ogbn-papers100m \cite{hu2020ogb} (see Table \ref{tab:dataset_statistics}). 
\revision{These datasets have varying average degrees, ranging from 5.6 to 60, and input feature sizes. They also differ on the fraction of vertices in the training set: in pubmed, this includes most vertices, whereas in papers it includes only 1.1\% of the vertices.}
Except for orkut, all these datasets come with input features that enable evaluating the accuracy of GNN training.
The orkut dataset contains only the graph topology, so we only use it to measure epoch time for the scalability experiments.
We consider the three standard GNN models used in previous work: GraphSAGE \cite{hamilton2017inductive}, GAT \cite{velikovi_graph}, and GCN \cite{kipf_semisupervised}.



\spara{Representative GNN training systems.} 
To compare across full-graph and mini-batch distributed GNN systems, we select representative systems that have publicly available and stable implementations and incorporate the main design choices discussed in Section~\ref{sec:mb-vs-fg-optimizations}, \revision{as summarized in Figure~\ref{fig:baselines}.}
We run all systems on top of PyTorch 2.0.1 and Python 3.8.10. 

\revision{
For full-graph training, in the training data management dimension of Figure~\ref{fig:baselines}, our evaluation only considers the more favorable situation where the entire training data is fully cached in the memory of the GPUs since they can entirely avoid the overhead of GPU-host communication.
All systems adopt a pipeline similar to the one of Figure~\ref{fig:fg_mb_training_diagram}(a).
The \textbf{Full-Graph} baseline is synchronous and does not use any optimization that can impact accuracy~\cite{pipegcngtihub_2023_pipegcn}.
In the dimension of algorithmic optimization, we consider three baselines representing three classes of algorithmic optimizations:
\textbf{PipeGCN} for asynchronous training~\cite{wan_pipegcn,pipegcngtihub_2023_pipegcn}, \textbf{BNS-GCN} for sampling~\cite{wan2022bnsgcn,bns-gcn-repo}, and \textbf{AdaQP} for message quantization~\cite{wan2023adaptive, adaqp-repo}.

For mini-batch training, we consider three baselines: \textbf{DGL}~\cite{wang2019dgl,dgl_2023_dmlcdgl}, its distributed version \textbf{DistDGL}~\cite{zheng2021distdgl,dgl_2023_dmlcdgl}, and \textbf{Quiver}~\cite{tan_quiver,quiver_2023_quiverteamtorchquiver}.
These cover the system-level optimizations discussed in Section~\ref{sec:mb-vs-fg-optimizations} (see Figure~\ref{fig:baselines}).
All systems adopt a pipeline similar to the one of Figure~\ref{fig:fg_mb_training_diagram}(b).
In terms of sampling implementations, DistDGL uses CPU-based sampling, while both DGL and Quiver support GPU-based sampling.
DistDGL partitions the training dataset across multiple hosts.
All systems use GPU caching whenever possible. 
DGL only supports local GPU caching while Quiver also supports distributed GPU caching, which is necessary to cache the Orkut and Papers100M datasets.
Mini-batch training can use different sampling algorithms to select a subset of the k-hop neighbors of the vertices in the mini-batch.
We consider the standard \textbf{Neighborhood Sampling (NS)} algorithm, which is available for both DGL and Quiver, \textbf{ClusterGCN}~\cite{chiang_2019_clustergcn}, and \textbf{GraphSaint}~\cite{graphsaint-ipdps19}.
}


\revision{
\spara{Hyperparameter search.} 
We run an extensive hyperparameter tuning.
We validated our search by matching the best test accuracies we found in the literature on GNN training systems we reviewed.
To manage the large search space, we iterate multiple times over three phases, where we tune a set of hyperparameters and fix all others until we don't observe any more improvements in accuracy.
In the first phase, we use grid search for the architectural hyperparameters, namely the number of layers or the hidden feature size, which have a limited number of discrete values. 
In the second phase, we run a random search for the remaining hyperparameters, such as the learning rate.
Mini-batch training requires tuning the micro-batch size and additional sampling hyperparameters compared to full-graph training.
We tune these in a third phase using Bayesian optimization, reducing the search space by considering only a few commonly used discrete values.
In the appendix, we report our hyperparameter space in table~\ref{parameter_search_space} and show that mini-batch training is less sensitive than full-graph training to the tuning of the architectural hyperparameters.
}




\section{Performance Evaluation}
\label{sec:perf_eval}

We evaluate the performance of our representative GNN training systems and answer two questions: \emph{Q1) How long do representative systems take to reach the same target accuracy?} (Section~\ref{sec:time_to_acc}), and \emph{Q2) How do these systems scale with a varying number of GPUs and hosts?} (Section~\ref{sec:scalability}).

\subsection{Time-to-Accuracy} \label{sec:time_to_acc}

\spara{Methodology}
\textit{Time-to-accuracy} is the time taken by any system to train to a target accuracy. 
To measure it, we select the target accuracies by leveraging hyperparameter tuning.
Given a dataset and a GNN model, we first find a model architecture on which both training approaches (full-graph and mini-batch) converge to a similar and close-to-best accuracy, which we call \textit{convergence accuracy}.
We select the \emph{target accuracy} as the minimum convergence accuracy across the two training approaches.
We then use the same model architecture across all GNN training systems and measure their time to reach the target accuracy.
\revision{The time-to-accuracy considers only the final training phase and not the hyperparameter search.}
For mini-batch systems, we consider the common Neighborhood Sampling algorithm as default because it is available on all baselines and it consistently achieves an accuracy close to the full-graph training methods. 
We report the hyperparameters in the 
appendix~\ref{sec:appendix}.
These experiments consider the two host types described in Section~\ref{sec:experimental_setup}: a higher-end \emph{NVLink} host and a lower-end \emph{PCIe} host.

\begingroup
\renewcommand{\arraystretch}{0.8}

\begin{table}[t]
\caption{Time to accuracy (TTA) and epoch time (ET) for GraphSAGE (NVLink host).}
\label{tab:time_to_acc_nvlink}
\tabcolsep=0.15cm 
    \begin{center}
    \begin{small}
\begin{tabular}{ |l|l*{6}{|r}|} 
\toprule
 & System & pubmed  & ogbn- & reddit & ogbn- & ogbn- \\
 &  &  & arxiv &  & products & papers100m \\
\midrule
\multirow{3}{1.5em}{ET (s)} 
& Full-Graph   & 0.009 & 0.057 & 1.845 & 2.086 & 1.094 \\
& PipeGCN  & 0.008 & \textbf{0.011} & \textbf{0.370} & 0.341 & \textbf{ 0.644} \\
& BNS-GCN   & 0.012 & 0.017 & 0.483 & 0.473 & 0.846 \\
& \revision{AdaQP}   & 0.027 & 0.029 & 0.830 & \textbf{0.312} & 0.471 \\ \cline{2-7} \noalign{\vskip 0.05cm}
& DGL       & 0.007 & 0.110 & 2.262 & 1.335 & 3.428 \\
& Quiver    & \textbf{0.005} & 0.131 & 1.994 & 2.117 & 1.878 \\
\midrule
\multirow{3}{1.5em}{TTA (s)} 
& Full-Graph  & 0.87 & 11.03 & 202.95 & 479.81 & 984.33 \\
& PipeGCN & 1.02 & 3.44 & 103.47 & 122.64 & 553.75 \\
& BNS-GCN  &  1.59 & 4.51 & 144.98 & 217.66 & 262.11\\ 
& \revision{AdaQP}   &  0.81 & 19.86 & 41.5 & 53.04 & 160.41\\ \cline{2-7} \noalign{\vskip 0.05cm}
& DGL      & 0.44 & \textbf{0.55} & 6.78 & \textbf{40.05}  & 75.42\\
& Quiver   & \textbf{0.33} & 0.65 & \textbf{5.98} & 63.52 & \textbf{41.31}\\
      \bottomrule
\end{tabular}
    \end{small}
    \end{center}
\end{table}

\endgroup

\begin{figure}
    \includegraphics[width=\linewidth]{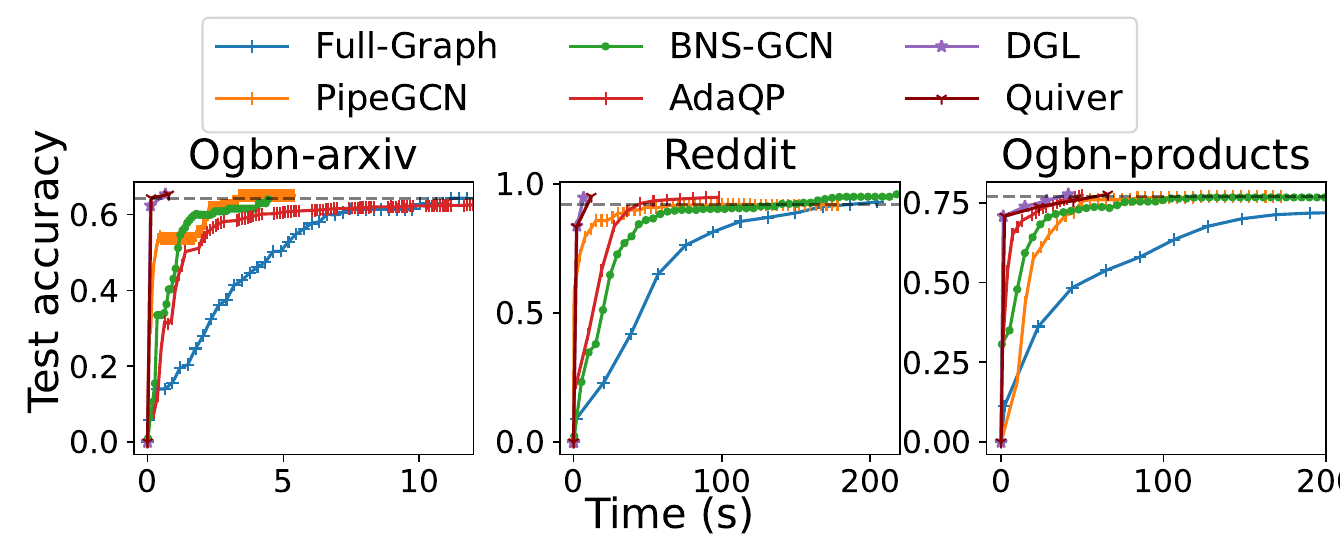}
    \caption{Convergence curve for GraphSAGE (NVLink host).}
    \label{fig:nvlink_time_to_acc}
\end{figure}

\spara{NVLink host.} \label{sec:time_to_acc_nvlink}
Table~\ref{tab:time_to_acc_nvlink} and Figure~\ref{fig:nvlink_time_to_acc} show the time to accuracy when training GraphSAGE on a single NVLink host with 4 GPUs. 
Full-graph training systems have a consistently lower epoch time (ET) than mini-batch training systems because they perform only one training iteration per epoch.
However,  their time-to-accuracy (TTA) is substantially higher compared to mini-batch systems. 
\revision{
For example, on the largest graph we consider, ogbn-papers100m, Quiver is $3.9\times$ faster than AdaQP. 
Similarly, for pubmed, Quiver is $2.5 \times$ faster than AdaQP; for arxiv, DGL is $6.3\times$ faster than PipeGCN; for reddit, Quiver is $6.9 \times$ faster than AdaQP (highest speed-up we observed); for products, DGL is 33\% faster than AdaQP (lowest speed-up we observed).}
Results on other models are shown in the 
appendix in table~\ref{tab:time_to_acc_nvlink_gat}.

\revision{Comparing the performance of systems in the same class shows the impact of the systems optimizations introduced in previous work (see Figure~\ref{fig:baselines})}. 
All full-graph systems we consider benefit from fully caching the graph in GPU memory.
PipeGCN can substantially decrease ET and TTA compared to Full-Graph. 
\revision{By using asynchronous training and overlapping communication with computation, PipeGCN can mitigate the cost of synchronous cross-GPU vertex feature communication at each layer, which is high even when NVLink.} 
Boundary-node sampling, as proposed by BNS-GCN, is less effective than asynchronous training on the smaller graphs, but it yields the fastest TTA among all full-graph training systems on ogbn-papers100M.
\revision{The use of of synchronous communication in BNS-GCN results in larger ET than PipeGCN for all datasets, but using fresh training data contributes to achieving a shorter number of epochs to convergence with papers100M.
AdaQP also combines synchrony with reduced communication costs thanks to message quantization.
It speeds up ET significantly compared to Full-Graph, but not compared to other full-graph optimizations.
Nonetheless, AdaQP can have faster TTA than other full-graph graph systems, since for some datasets it requires fewer epochs to converge.

Among mini-batch systems, DGL supports local full caching and can cache all the graphs in Table~\ref{tab:time_to_acc_nvlink} except papers100m.
When full local caching is possible, DGL does not need to transfer input vertex features to the GPUs, maximizing GPU utilization.
Quiver also uses full local caching, achieving a similar ET and TTA.
However, Quiver implements a partial distributed cache across multiple GPUs that can also cache ogbn-papers100m.
With distributed caching, some input vertex features may still be transferred among GPUs using NVLink buses, which is cheaper than loading the features from the host memory over the slower PCIe bus.
This is why Quiver has a much lower time-to-accuracy than DGL for papers100M.
}

\begingroup
\renewcommand{\arraystretch}{0.8}
\begin{table}[t]
\caption{Time to accuracy (TTA) and epoch time (ET) for GraphSAGE (PCIe host).}
\label{tab:time_to_acc}
\tabcolsep=0.135cm 
    \begin{center}
    \begin{small}
\begin{tabular}{ |m{2.8em} |m{9em}  |>{\raggedleft\arraybackslash}m{3.0em} |>{\raggedleft\arraybackslash}m{2.5em} | >{\raggedleft\arraybackslash}m{2.5em} |>{\raggedleft\arraybackslash}m{3.5em}|} 
\toprule
 & System & pubmed  & ogbn-arxiv & reddit & ogbn-products \\
\midrule

\multirow{3}{6em}{ET (s)} 
& Full-Graph & 0.024 & 0.188 & 5.602 & 5.835  \\
& PipeGCN & 0.020 &\textbf{ 0.067} & \textbf{3.278} & \textbf{2.526} \\
& BNS-GCN & 0.031 & 0.072 & 3.399 & 2.915  \\
\cline{2-6} \noalign{\vskip 0.05cm}
& DGL & \textbf{0.019} & 0.314 & 9.416 & 7.081 \\
& Quiver & \textbf{0.019} & 0.216 & 10.272 & 9.246  \\
\midrule

\multirow{3}{5em}{TTA (s)} 
& Full-Graph & 2.40 & 36.16 & 616.25 & 1341.95  \\
& PipeGCN & 2.35 & 20.75 & 917.80 & 909.43 \\
& BNS-GCN & 2.91 & 19.33 & 1019.64 & 1340.80  \\
\cline{2-6} \noalign{\vskip 0.05cm}
& DGL & 1.25 & 1.57 & \textbf{28.25} & \textbf{247.83} \\
& Quiver & \textbf{1.22} & \textbf{1.08} & 30.82 & 323.61 \\
      \bottomrule
\end{tabular}
    \end{small}
    \end{center}
\end{table}
\endgroup

\spara{PCIe host.} 
\label{time_to_acc_pcie} 
Table~\ref{tab:time_to_acc} reports the results for single-host training with GraphSAGE using a lower-end host setup, where NVLink is not available and all cross-GPU communication must occur over a slower PCIe bus.
The 4 GPUs are computationally less powerful and have less memory.
The papers100M graph is too large to be fully cached in a single host by our full-graph training systems.
\revision{
We cannot run AdaQP since its implementation is not compatible with our lower-end GPUs.

Like in the previous hardware setups, mini-batch training systems show a larger ET (with the exception of pubmed) but a shorter TTA on all datasets.
Full-graph training systems must run cross-GPU vertex feature exchange over the slow PCIe bus.
Optimizations that reduce the communication cost, like asynchrony (PipeGCN) and sampling (BNS-GCN), can reduce TTA, but the cost remains high.
DGL and Quiver avoid vertex feature communication by replicating the training data on all GPUs so the advantage over the full-graph training systems is larger than with the NVLink host, reaching $32\times$ for Reddit.
}

\revision{
\spara{The impact of dataset characteristics.}
Different datasets have an impact on the performance of different systems.
While the impact on TTA is hard to understand due to the stochastic nature of training, the impact on ET can be more easily analyzed.
}

\revision{The communication cost of full-graph training systems is determined by the number of boundary vertices across partitions.
Reddit and products, for example, have a large number of boundary vertices, so the ET of the Full-Graph baseline is much larger compared to DGL and Quiver.
Optimizations that reduce this communication cost used by PipeGCN and BNS-GCN can reduce this gap significantly.
For mini-batch training, the training cost is determined by the size of the micro-batches assigned to each GPU. 
Datasets that require a larger number of GNN layers, such as products and reddit which requires 5 and 4 layers respectively, result in a larger ET for mini-batch training systems compared to the other full-graph training systems.
In contrast, papers100M only requires 2 layers so the gap in ET is smaller.
}

\revision{
\spara{Different mini-batch sampling algorithms.}
Finally, we combine DGL with different mini-batch sampling algorithms than Neighborhood Sampling (NS). ClusterGCN achieves on average $2.8\times$ faster TTA than NS across datasets, ranging from $0.14\times$ slower to $12.3\times$ faster depending on the dataset, and its ET is $6.4\times$ faster on average.
GraphSaint has $3.2\times$ faster TTA than NS on average, ranging from $0.5\times $ slower to $16.1\times$, and faster ET by $9.3\times$ on average 
(see appendix tables~\ref{tab:time_to_acc_gat_samplers} and \ref{tab:time_to_acc_gcn_samplers}).
}

\spara{Takeaway.} Mini-batch training systems converge faster than full-graph training ones because they require fewer epochs to reach a target accuracy, despite having longer epochs. 
We observe this across all models, datasets, and hardware configurations we consider.
\revision{Avoiding data transfers by caching input data in GPU memory has a strong impact on the performance of mini-batch training. 
For full-graph training, asynchrony (PipeGCN) and sampling (BNS-GCN) can reduce communication cost and thus ET, resulting in faster TTA.
AdaQP can achieve even faster TTA even if its ET is larger than other full-graph methods, showing the importance of using algorithmic optimizations besides system-level optimizations.
For mini-batch training, using different sampling algorithms can speed up TTA significantly in many cases.
}


\subsection{Scalability} \label{sec:scalability}

\begingroup
\renewcommand{\arraystretch}{0.7}
\begin{table*}[t]
\caption{Single-host scalability for GAT (NVLink host).  ET - Epoch time (s), TTA - Time to Accuracy (s).}
\label{tab:scalability_gat_a100_single_host}
    \begin{center}
    \begin{small}
\begin{tabular}{|m{7em} | m{4em}| >{\raggedleft\arraybackslash}m{2.5em} >{\raggedleft\arraybackslash}m{2.5em} >{\raggedleft\arraybackslash}m{4em} | >{\raggedleft\arraybackslash}m{2.5em} >{\raggedleft\arraybackslash}m{3em} >{\raggedleft\arraybackslash}m{4em}| >{\raggedleft\arraybackslash}m{2.5em} >{\raggedleft\arraybackslash}m{3em} >{\raggedleft\arraybackslash}m{4em}| >{\raggedleft\arraybackslash}m{2.5em}  >{\raggedleft\arraybackslash}m{4em} |}
\toprule
Dataset &  & \multicolumn{3}{c|}{ogbn-arxiv} & \multicolumn{3}{c|}{reddit} & \multicolumn{3}{c|}{ogbn-products} &  \multicolumn{2}{c|}{orkut}  \\
\midrule 
System & \# of GPUs & ET & TTA & TTA Speed-up & ET & TTA  & TTA Speed-up & ET & TTA  & TTA Speed-up & ET & ET speed-up \\
\midrule 
\multirow{4}{*}{Full-Graph} 
& 1 & 1.293 & 569.01 & 1.0x & 4.265 & 4094.59 & 1.0x & 2.058 & 823.36 & 1.0x & 1.898 & 1.0x\\
& 2 &0.868 & 381.74 & 1.5x & 2.566 & 2463.65 & 1.7x & 1.341 & 536.20 & 1.5x & 1.873 & 1.0x\\
& 3 & 0.763 & 335.90 & 1.7x & 2.587 & 2483.90 & 1.6x & 1.129 & 451.64 & 1.8x & 1.594 & 1.2x\\
& 4 & 0.625 & 275.04 & 2.1x & 1.801 & 1728.58 & 2.4x & 1.033 & 413.28 & 2.0x & 1.442& 1.3x\\
\midrule
\multirow{4}{*}{PipeGCN} 
& 1 & 1.292 & 581.27 & 1.0x & 4.268 & 4267.70 & 1.0x & 2.060 & 782.76 & 1.0x & 1.898 &  1.0x\\
& 2 & 0.777 & 349.83 & 1.7x & 2.309 & 2309.20 & 1.8x & 1.135 & 431.34 & 1.8x & 1.149 & 1.7x\\
& 3 & 0.589 & 265.10 & 2.2x & 2.309 & 2309.20 & 1.8x & 0.810 & 307.61 & 2.5x & 0.920 & 2.1x\\
& 4 & 0.469 & 211.14 & 2.8x & 1.312 & 1312.15 & 3.3x & 0.614 & 233.24 & 3.4x & 0.720& 2.6x\\
\midrule
\multirow{4}{*}{BNS-GCN} 
& 1 & 1.309 & 471.06 & 1.0x & 4.189 & 4188.60 & 1.0x & 2.171 & 455.95 & 1.0x & 2.173 &  1.0x\\
& 2 & 0.764 & 274.93 & 1.7x & 2.315 & 2315.30 & 1.8x & 1.213 & 254.73 & 1.8x & 1.284 & 1.7x\\
& 3 & 0.532 & 191.56 & 2.5x & 1.724 & 1723.60 & 2.4x & 0.857 & 179.99 & 2.5x & 0.984 & 2.2x\\
& 4 & 0.420 & 151.20 & 3.1x & 1.395 & 1395.40 & 3.0x & 0.629 & 132.13 & 3.5x & 0.713 & 3.0x\\
\hline 
\midrule
\multirow{4}{*}{DGL} 
& 1 &15.006 & 150.06 & 1.0x & 24.006 & 360.09 & 1.0x & 10.554 & 158.31 & 1.0x & 14.791 & 1.0x\\
& 2 & 7.290 & 72.90 & 2.1x & 12.794 & 191.91 & 1.9x & 5.827 & 87.40 & 1.8x & 7.668 & 1.9x\\
& 3 & 5.550 & 61.04 & 2.5x & 8.949 & 134.24 & 2.7x & 3.894 & 58.42 & 2.7x & 4.940 & 3.0x\\
& 4 & 3.871 & 42.58 & 3.5x & 6.510 & 97.65 & 3.7x & 2.730 & 40.95 & 3.9x & 3.980 & 3.7x\\
\midrule
\multirow{4}{*}{Quiver} 
& 1 & 13.813 & 138.13 & 1.0x & 26.427 & 396.41 & 1.0x & 11.212 & 168.19 & 1.0x & 13.718 & 1.0x\\
& 2 & 6.575 & 65.75 & 2.1x & 13.764 & 206.46 & 1.9x & 5.797 & 86.95 & 1.9x & 7.447  & 1.8x\\
& 3 & 4.979 & 49.79 & 2.8x & 8.703 & 130.55 & 3.0x & 3.993 & 59.90 & 2.8x & 5.058  & 2.7x\\
& 4 & 3.596 & 35.96 & 3.8x & 6.779 & 101.68 & 3.9x & 2.974 & 44.61 & 3.8x & 3.855 & 3.6x\\
\bottomrule
\end{tabular}
    \end{small}
    \end{center}
    \vskip -0.1in
\end{table*}
\endgroup
\begingroup
\renewcommand{\arraystretch}{0.7}
\begin{table}[t]
\caption{Distributed scalability for GraphSage (NVLink hosts). ET - Epoch time (s), TTA - Time to Accuracy (s).}
\label{tab:dist_scalability_graphsage_nvlink}
    \begin{center}
    \begin{small}
    \tabcolsep=0.09cm 
\begin{tabular}{|l | l | >{\raggedleft\arraybackslash}m{2.5em} >{\raggedleft\arraybackslash}m{4em} | >{\raggedleft\arraybackslash}m{2.5em} >{\raggedleft\arraybackslash}m{2.5em} >{\raggedleft\arraybackslash}m{4em}|}
\toprule
 &  & \multicolumn{2}{c|}{orkut} & \multicolumn{3}{c|}{ogbn-papers100m} \\
\midrule 
System& \# Hosts & ET & ET Speed-up & ET & TTA & TTA Speed-up \\
\midrule 
\multirow{2}{*}{Full-Graph} 
& 1 & 4.769 & 1.0x & 3.556 & 924.46& 1.0x\\
\multirow{2}{*}{}& 2 & 4.588 & 1.0x & 3.566 & 927.13 & 1.0x\\
& 3 & 4.351 & 1.0x & 3.933 & 1022.63 & 0.9x  \\
\midrule
\multirow{2}{*}{PipeGCN} 
& 1 & 0.454 & 1.0x & 0.759 & 197.57 & 1.0x\\
\multirow{2}{*}{} & 2 & 0.341 & 1.3x & 0.464 & 120.67 & 1.6x \\
& 3 & 0.227 & 2.0x & 0.361 & 93.96 & 2.1x \\
\midrule

\multirow{2}{*}{BNS-GCN} 
& 1 &0.810 & 1.0x & 0.997 & 209.38 & 1.0x \\
\multirow{2}{*}{}& 2 & 0.608 & 1.3x  & 0.762 & 160.10 & 1.3x \\
& 3 & 0.405 & 2.0x & 0.542 & 113.88 & 1.8x \\
\hline 
\midrule
\multirow{2}{*}{DistDGL} & 1 & 9.83 & 1.0x & 11.56 & 80.93 & 1.0x \\
\multirow{2}{*}{}& 2 & 7.31 & 1.3x & 10.75 & 75.24 & 1.1x \\
& 3 & 6.82 & 2.0x & 6.61 & 46.28 & 1.7x \\
\midrule
\multirow{2}{*}{\revision{Quiver}} & 1 & 7.33 & 1.0x & 2.63 & 18.41 & 1.0x \\
\multirow{2}{*}{}& 2 & 4.67 & 1.6x & 1.96 & 13.72 & 1.3x \\
& 3 & 3.78 & 1.9x & 1.66 & 11.62 & 1.6x \\
\bottomrule
\end{tabular}
    \end{small}
    \end{center}
    \vskip -0.2in
\end{table}

\endgroup

Next, we answer \emph{Q2: How do systems scale with a varying number of GPUs and a varying number of hosts?}
For each dataset, we pick a GNN architecture that fits in a single GPU and use the single GPU training performance to calculate the scalability speedups.

\spara{Single-host scalability (NVLink).} \label{sec:pcie_scalability_single_host}
We now measure single-host scalability by varying the number of GPUs.
We report the hyperparameters in the 
appendix in table~\ref{tab:scalability_architectures_single_host}
For the orkut dataset, we do not report TTA since it has synthetic input features and target labels. 

Table \ref{tab:scalability_gat_a100_single_host}
presents the results for GAT models. 
Like we observed previously, the full-graph training systems have a much lower ET than the mini-batch ones but a higher TTA in all configurations.

Mini-batch systems have a lower TTA than full-graph ones also in a single GPU setting, indicating that they have a lower computation cost to start with. 
The mini-batch systems scale almost linearly, up to $3.9\times$. 
When scaling to multiple GPUs, one important advantage of mini-batch training systems is that they can keep the amount of work per GPU constant by increasing the mini-batch size.
This is not possible in full-graph training systems since the total amount of work per epoch is constant.
As the number of GPUs increases, the amount of work per GPU decreases but the communication cost increases, hindering scalability.
This explains why full-graph training systems that optimize communication scale better than the Full-Graph baseline.
PipeGCN has much better scalability because it overlaps communication and computation using asynchronous training, scaling to up to $3.4\times$. 
Sampling of boundary nodes in BNS-GCN also enables good scaling, up to $3.5\times$, since the cross-GPU communication is avoided.


Results on GraphSAGE models are shown in the 
appendix in table~\ref{tab:scalability_graphsage_a100_single_host}
The Full-Graph baseline does not scale at all, PipeGCN scales up to $3.3\times$ and BNS-GCN scales up to $2.7\times$. 
In contrast, DGL and Quiver are able to scale up to $\approx 4.0\times$ and $3.8\times$ respectively.

\spara{Multi-host scalability (NVLink).} \label{sec:nvlink_scalability_multi_host}
We now consider scaling the training of larger graphs (orkut and ogbn-papers100m) in a distributed setting of NVLink hosts, where each host has 4 GPUs.
For this experiment, we replace DGL with DistDGL, its distributed version. We report the hyperparameters in the
in the appendix table~\ref{tab:scalability_architectures_multi_host}.
We consider the GraphSAGE model because it is the only one implemented in the DistDGL distribution. 
\revision{We run Quiver with data parallelism across multiple hosts.}

\begin{figure}
    \centering
    \includegraphics[width=0.6\columnwidth]{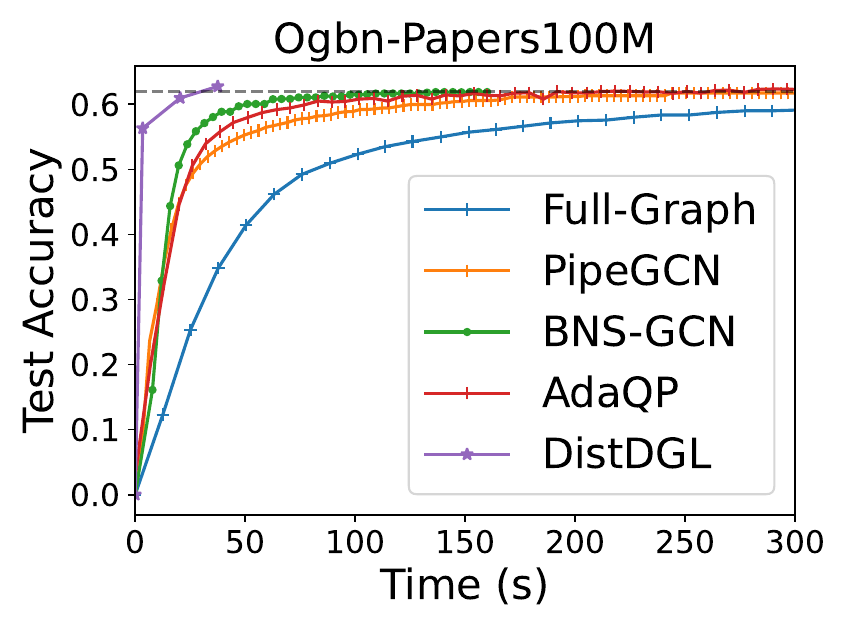}
    \vskip -0.1in
    \caption{Convergence curve for GraphSage (3 NVLink hosts).}
    \label{fig:nvlink_papers_time_to_acc}
    \vskip -0.2in
\end{figure}

Table \ref{tab:dist_scalability_graphsage_nvlink} and Figure \ref{fig:nvlink_papers_time_to_acc} show the results for distributed scalability with NVLink-enabled devices. 
Also, in this case, we observe that full-graph training systems consistently have lower epoch time but larger time-to-accuracy compared to DistDGL.
\revision{
Going from single-host to multi-host training impacts different systems in a different way.
Full-graph training systems must use the network to exchange vertex features across hosts, which is much slower than NVLink.
The Full-Graph baseline actually has a slower epoch time than a single-host setup.
Optimizing communication using asynchrony (PipeGCN) and boundary node sampling (BNS-GCN) is even more crucial to scaling than in the single-host scalability experiments.
DistDGL partitions the dataset across multiple hosts and performs distributed sampling, which introduces an additional communication cost compared to a single-host implementation. }
\revision{When running on multiple hosts, the DistDGL baseline has a much larger epoch time than Quiver because it uses distributed CPU-based sampling and does not use caching. }


\spara{Scalability with the PCIe host.} \label{sec:pcie_scalability_multi_host} 
We run single-host scalability experiments on the PCIe host, from 1 to 4 GPUs.
Like for the previous experiments, mini-batch training systems have a lower time-to-accuracy than full-graph training ones despite having a larger epoch time.
The results show that all the full-graph training systems have lower scalability since they must not communicate over the slower PCIe bus instead of NVLink.
This slowdown compared to the NVLink case is particularly evident for the Full-Graph baseline, which can only scale up to $2.0\times$.
PipeGCN async can scale up to at most $3.1\times$ and BNS-GCN up to at most $3.0\times$.
Mini-batch training systems have better scalability because they replicate most graphs in the cache and require less cross-GPU communication.
DGL scales up to $3.8\times$ and Quiver up to $3.5\times$.

We also run all systems using 3 PCIe hosts for the ogbn-papers100m graph. 
Compared to the distributed NVLink setup, hosts are connected by a much slower network (1 Gbps instead of 25 Gbps).
Like in the previous experiments, DistDGL has a lower time-to-accuracy than the other full-graph training systems, but the difference is smaller than with the distributed NVLink setup because of the cost of distributed sampling and feature loading over a very slow network.
DistDGL outperforms the BNS-DGL system by only $12\%$.
The distributed sampling implementation of DistDGL is not optimized for very slow network links.
The full results are shown in the 
appendix in table~\ref{tab:scalability_gat_m40_single_host}.

\revision{
\spara{Memory scalability.}
In terms of memory usage, full-graph training systems must keep the entire graph and the related training state partitioned in GPU memory.
Mini-batch systems only need to store a micro-batch and the related state in the memory of each GPU.
In our experiments, full-graph training systems have a larger peak memory utilization in all the configurations we considered, up to $7.4\times$, but the gap decreases when using more GPUs (since mini-batch training systems must store more micro-batches in total), deeper GNNs with more layers, or larger sampling fanouts (since each micro-batch is larger).

}

\spara{Takeaway.} 
Mini-batch training systems have lower time-to-accuracy than full-graph training ones for all models, datasets, and hardware configurations we consider.
They also have better scalability in most cases. 
One important scalability advantage for mini-batch training systems is that they can tune the mini-batch size to keep the work per GPU constant.
In full-graph training systems, algorithmic optimizations such as asynchrony and sampling are essential to achieve high scalability.
The full-graph systems cannot scale well when they use slower cross GPU interconnects because their communication cost is amplified.


\section{Accuracy Evaluation} \label{sec:results_section}


Even though a system takes longer than another to reach the \emph{same} target test accuracy, it could still be preferable to use it if it can converge to a \emph{higher} test accuracy.
Indeed, some works on GNN systems assert that full-graph training achieves higher accuracy and that mini-batch training does not guarantee convergence \cite{thorpe_dorylus, wan_pipegcn, wan2022bnsgcn, G3_full_graph, DBLP_tripathy_cagnet, cai_2021_dgcl, yang_2023_betty, wan2023adaptive, sancus, mostafa2022sequential, ma_neugraph}, while others claim that mini-batch training achieves higher accuracy \cite{kaler2022accelerating_salient, zou_2019_layerdependent, zheng_2022_distributed, cm_gcn_zhao, zheng2021distdgl}. 

\begin{figure}
    \centering
    \includegraphics[width=0.45\textwidth]{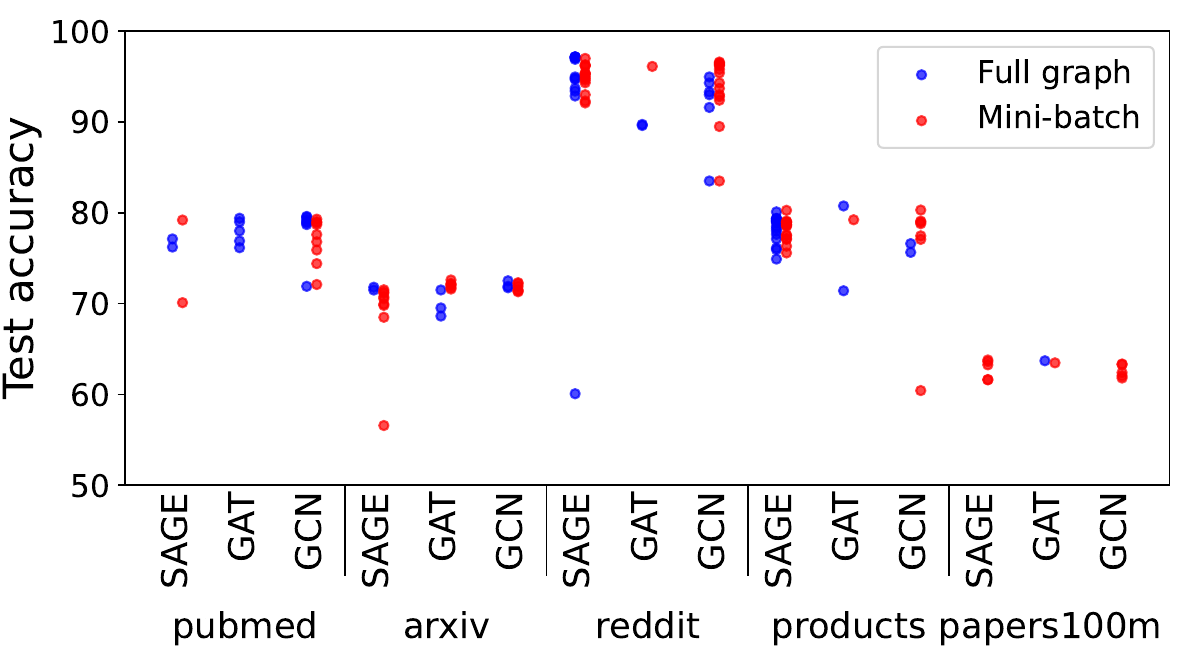}
    \caption{Test accuracy reported in literature on GNN training~\cite{wan_pipegcn, dgl_2023_dmlcdgl, hu2020ogb, song_2021_network, md2021distgnn, yang_2023_betty, li2024rethinking, pmlr-v198-gasteiger22a_influence_based, yao2021blocking, zou_2019_layerdependent, hamilton2017inductive, chiang_2019_clustergcn, graphsaint-ipdps19, chen2018stochastic_vrgcn, yao2021blocking, wu2024sgformer, velikovi_graph, thorpe_dorylus, frasca2020sign, wan2022bnsgcn, kipf_semisupervised, jia_improving_roc}
    .}
    \label{fig:literature_acc}
    \vskip -0.1in
\end{figure}

Figure~\ref{fig:literature_acc} summarizes the test accuracy values reported in the literature we reviewed for different GNN models and datasets.
Each point in the graph represents the test accuracy reported for a specific GNN model-dataset pair in one of the papers we reviewed. 
The figure illustrates a wide distribution of accuracy values for the same dataset and model combination. 

In this Section, we shed light onto this apparent contradiction and answer two questions: \emph{Q3) What is the best test accuracy we can achieve using full-graph and mini-batch training? Q4) What test accuracy can we achieve when using a training method with the best hyperparameters found for the other method?}
\revision{Different training systems have different optimizations (\finalrevision{see Figure~\ref{fig:baselines})}, so \emph{Q5: What is the impact of the optimizations on accuracy?} 
}

\begin{figure}
    \centering
    \includegraphics[width=0.45\textwidth]{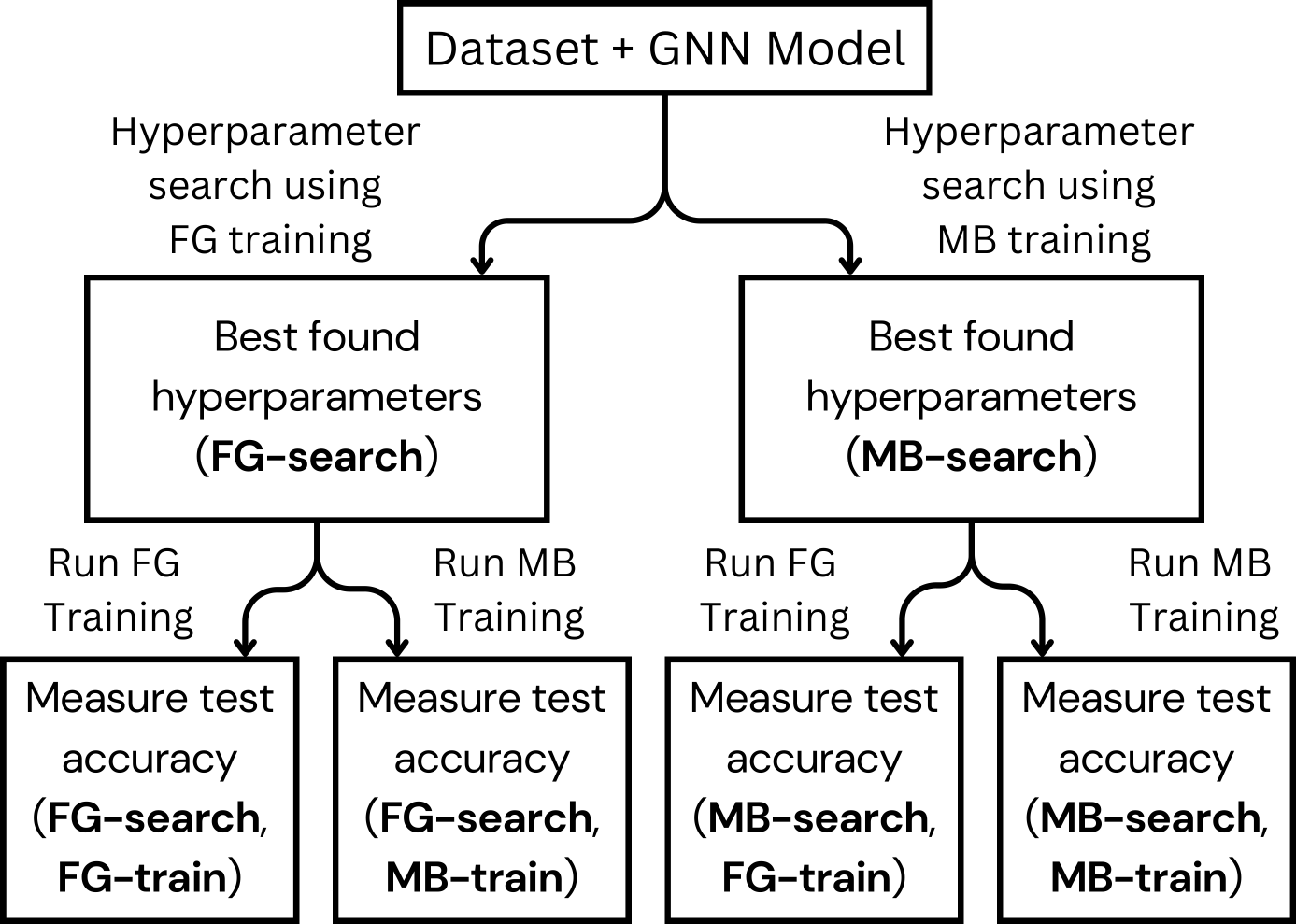}
    \caption{Methodology of our accuracy comparison between full-graph (FG) and mini-batch (MB) training.}
    \label{fig:methodology}
\end{figure}

\spara{Methodology.}
To compare maximum test accuracies among full-graph and mini-batch training, we use the methodology depicted in Figure~\ref{fig:methodology}.
Given a dataset and a GNN model, we conduct two separate hyperparameter searches, one using full-graph training and one using mini-batch training.
This is because the two training methods can have different sets of best hyperparameters for the same dataset and model, using the method described in Section~\ref{sec:experimental_setup}.
We consider \emph{vanilla} baselines for full-graph and mini-batch training: the Full-Graph baseline and DGL with the default Neighborhood Sampling algorithm, respectively.
The best hyperparameter settings we obtained using the two training methods are denoted as \textsc{FG-search} and \textsc{MB-search}.
We then run full-graph and mini-batch training on both settings, measure the test accuracy, and denote it as \textsc{FG-train} and \textsc{MB-train} respectively.
This yields a total of four test accuracy combinations per dataset and GNN model.
We report the hyperparameter settings we found in the 
appendix in table~\ref{tab:best_architectures}.

We validate our results by verifying that given a dataset, model, and training method, our hyperparameter search achieves accuracy values that are equal to or better than the best accuracy value reported in the literature reviewed in Section~\ref{sec:review}.


\begin{table*}[t]
\caption{Accuracy (\%) of GraphSage using hyperparameter tuning method in Figure~\ref{fig:methodology}.}
\label{tab:convergence_graphsage}
    \begin{center}
    \begin{small}
    \tabcolsep=0.035cm
\begin{tabular}{|l | l l | l l | l l  |l l |l l|} 
\toprule
 & \multicolumn{2}{c|}{Pubmed}  & \multicolumn{2}{c|}{Ogbn-arxiv} & \multicolumn{2}{c|}{Reddit} & \multicolumn{2}{c|}{Ogbn-products} & \multicolumn{2}{c|}{Ogbn-papers100M}\\
\midrule
& FG-search & MB-search & FG-search & MB-search & FG-search & MB-search & FG-search & MB-search & FG-search & MB-search \\
\midrule
FG-train & $\textbf{77.40} \pm \textbf{00.04}$ & $76.10 \pm 00.67$ & $\textbf{71.76} \pm \textbf{00.16}$ & $67.21 \pm 00.9$ & $\textbf{96.94} \pm \textbf{00.01}$ & $90.28 \pm 00.63$ & $\textbf{79.29} \pm \textbf{00.15}$ & $76.60 \pm 00.32$ & $\textbf{62.82} \pm \textbf{00.01}$ & $61.74 \pm 00.13$ \\
MB-train & $77.00 \pm 00.39$ & $\textbf{78.20} \pm \textbf{00.32}$ & $70.73 \pm 00.38$ & $\textbf{71.67} \pm \textbf{00.30}$ & $96.52 \pm 00.14$ & $\textbf{96.82} \pm \textbf{00.04}$ & $78.72 \pm 00.07$ & $\textbf{78.84} \pm \textbf{00.12}$ & $62.52 \pm 00.04$  & $\textbf{62.77} \pm \textbf{00.02}$\\
      \bottomrule
\end{tabular}
    \end{small}
    \end{center}
\end{table*}

\begingroup
\renewcommand{\arraystretch}{0.8}

\begin{table}[t]
\caption{\revision{Accuracy (\%) impact of different optimizations - GraphSAGE.}}
\label{tab:accuracies_sage_all}
    \begin{center}
    \begin{small}
    \revision{
\begin{tabular}{|l | >{\raggedleft\arraybackslash}m{2.5em} >{\raggedleft\arraybackslash}m{4em} >{\raggedleft\arraybackslash}m{2.5em} >{\raggedleft\arraybackslash}m{3em} >{\raggedleft\arraybackslash}m{4.5em}|
}
\toprule
& Pubmed & Ogbn-arxiv & Reddit & Ogbn-products & Ogbn-papers100M \\
\midrule
Full-Graph & $77.40 \pm 0.04$ & $71.76 \pm 0.16$ & $96.94 \pm 0.01$ & $79.29 \pm 0.15$ & $62.82 \pm 0.01$ \\
PipeGCN & $77.40 \pm 0.45$ & $71.27 \pm 0.66$ & $96.93 \pm 0.11$ & $78.58 \pm 0.32$ & $61.74 \pm 0.13$ \\
BNS-GCN & $77.20 \pm 0.16$ & $71.72 \pm 0.27$ & $96.95 \pm 0.08$  & $78.40 \pm 0.22$ & $62.79 \pm 0.10$ \\
AdaQP & $77.42 \pm 0.59$ & $63.73 \pm 0.07$ & $96.85 \pm 0.002$  & $78.79 \pm 0.004$ & $62.54 \pm 0.00$ \\
\hline \hline 
NS & $78.20 \pm 0.32$ &$71.67 \pm 0.30$ & $96.82 \pm 0.04$ & $78.84 \pm 0.12$ & $62.77 \pm 0.02$\\
ClusterGCN & $79.72 \pm 0.22$ & $65.72 \pm 0.18 $ & $95.07 \pm 0.03$ & $79.50 \pm 0.17$ & $49.79 \pm 0.00$ \\ 
GraphSaint & $84.03 \pm 0.61$  & $73.80 \pm 0.13 $ & $98.40 \pm 0.03$ & $77.85 \pm 0.07 $ & $50.32 \pm 0.27$ \\ 

\bottomrule
\end{tabular}
}
    \end{small}
    \end{center}
    \vskip -0.1in
\end{table}
\endgroup




\spara{Accuracy with the best setting.}
We start by answering question \emph{Q3:  What is the best test accuracy we can achieve using vanilla full-graph and mini-batch training? }
The accuracy results for the GraphSAGE model are shown in Table~\ref{tab:convergence_graphsage}. 
Given a dataset, we must compare the accuracies obtained for two combinations: (\textsc{FG-search}, \textsc{FG-train}) for full-graph training and (\textsc{MB-search}, \textsc{MB-train}) for mini-batch training. The results are averaged after 5 runs.

The results show that the accuracies are similar across training methods.
For the GraphSAGE model, each method can yield higher accuracy than the other, depending on the dataset.
\revision{Accuracy tends to have lower variance with full-graph training generally than with mini-batch training.}
The largest gap between the accuracies obtained with the two methods is less than $0.9\%$ in favor of full-graph training and averaging around $0.3\%$.
We ran a similar analysis for the GCN and GAT model, and we report the results in the 
appendix in table~\ref{tab:accuracies_gat_all} and table~\ref{tab:accuracies_gcn_all}.
With the GAT model, the largest gap is slightly higher, up to $1.1\%$ in favor of full-graph training and the average is around $0.6\%$. 
Gaps are higher for GCN, which is a simpler model and turns out to be more sensitive to the choice of hyperparameters. 
The largest gap in this case is $3.6\%$ in favor of mini-batch training and the average is around $1.27\%$. 



\spara{Accuracy with sub-optimal setting.}
We now answer question \emph{Q4: What test accuracy can we achieve when using a vanilla training method with the best hyperparameters found for the other method?}

To answer this question for the case of the GraphSAGE model, we still look at Table~\ref{tab:convergence_graphsage}.
Given a dataset, the \textsc{FG-search} column reports the accuracy achieved with the two training methods using the hyperparameter setting where full-graph training works best.
Mini-batch training performs worse than full-graph training, but only by a small margin. The largest gap between \textsc{MB-train} and \textsc{FG-train} is for the ogbn-arxiv graph ($1.03\%$).

The \textsc{MB-search} column compares the two training methods using the best hyperparameters we found for mini-batch training.
Full-graph training has lower accuracy than mini-batch training and the gap between the two methods is larger, reaching $6.22\%$ for the reddit dataset.

Our evaluation of GAT and GCN is reported in the
appendix in table~\ref{tab:convergence_gat} and table~\ref{tab:convergence_gcn}
which leads to similar conclusions. For GAT, the largest difference between FG-train and MB-train is $2.85\%$ in favor of MB-search for arxiv and $1.9\%$ in favor of FG-search for pubmed. For GCN, it is $1.36\%$ in favor of FG-search for reddit and $5.5\%$ in favor of MB-search for pubmed. 

These results motivate the use of separate hyperparameter tuning for the two methods since good hyperparameter settings do not transfer well across methods.

\revision{
\spara{Impact of optimizations on accuracy.}
\label{sec:impact-accuracy}
Many performance optimizations we consider can result in a different accuracy than the vanilla methods \finalrevision{(see Figure~\ref{fig:baselines})}.
This section answers the question \emph{Q5: What is the impact of the optimizations on accuracy?}
In this analysis, we use the best-found hyperparameters for each system class: \textsc{FG-Search} and \textsc{MB-Search} models for full-graph and mini-batch training systems, respectively.
For full-graph training, we consider the impact of asynchrony (PipeGCN), sampling (BNS-GCN), and message quantization (AdaQP).
For mini-batch training, besides Neighborhood Sampling (NS), we also consider the ClusterGCN and GraphSaint sampling algorithms.

The results for GraphSage are shown in Table~\ref{tab:accuracies_sage_all}.
Among full-graph training systems, our results show that asynchrony and sampling have minimal impact on accuracy compared to FG, within 1\% in all the cases we considered.
Message quantization has a stronger negative impact only on arxiv.

Mini-batch training shows a very different trend.
For most datasets, there is some sampling algorithm that achieves higher accuracy than all full-graph training approaches.
The relative performance of each sampling algorithm shows large variations depending on the dataset.
Both GraphSaint and ClusterGCN achieve the best and the worst accuracy among all sampling algorithms for some datasets.
These results show that the regularization effect of sampling is key to achieving high accuracy, focusing the training on nodes that have the highest influence on each other~\cite{graphsaint-ipdps19}.
Neighborhood Sampling (NS) shows a more stable behavior: on all datasets, it achieves similar accuracy as FG.
This justifies its popularity as the default sampling method for mini-batch training.
We repeated the experiment on GAT and GCN and obtained similar trends, as shown in the 
appendix in table~\ref{tab:accuracies_gat_all} and table~\ref{tab:accuracies_gcn_all}.
}


\spara{Takeaway.}
When comparing vanilla full-graph and mini-batch training, no method consistently guarantees higher accuracy than the other and the gap between the two methods is \revision{not larger than 3.6\%}. 
Good hyperparameter settings do not perform as well across both full-graph and mini-batch training, motivating the need for separate tuning processes.
\revision{
The algorithmic optimizations for full-graph training we considered achieve very similar accuracies as the vanilla full-graph method in most cases, making them attractive choices given their performance benefits.
For mini-batch training, using the right sampling algorithm can achieve higher accuracy than the full-graph training methods for most datasets and models. 
However, some sampling algorithms show widely different accuracies based on the dataset and sometimes converge to much lower accuracies.
This indicates that trying different algorithms during hyperparameter tuning is essential.

Overall, these results challenge the common rationale for using full-graph training over mini-batch training, suggesting that the expected benefits of avoiding sampling and its associated information loss to achieve higher accuracy do not materialize in practice.
}
\revision{\section{Cost Analysis}
\label{sec:analysis}
To support and generalize our empirical observations beyond the specific hardware configurations and software implementations we evaluated empirically, in this section we answer the following question \emph{Q6: What are the analytical performance costs of vanilla full-graph and mini-batch training?}
We model the communication and computation costs of \emph{training} analytically and then evaluate the cost of \emph{pre-processing} (sampling) in mini-batch training experimentally to complement the analysis.

\spara{Communication cost.}
Our analysis considers a simplified model of GNN training where the training workload is partitioned among a set of workers $W$, each accessing a local memory.
The communication cost is the volume of vertex feature data exchanged among the workers.
We ignore the communication cost of gradient synchronization since GNN models have relatively few parameters.
We measure the communication cost to convergence rather than the cost per epoch to account following the discussion of Section~\ref{sec:time_to_acc}.

In full-graph training, at each layer, workers must receive the features of all the vertices in other partitions that have neighbors in the local partition. 
The communication cost is:
$$
\Gamma_{\textrm{fg}} = n_f \sum_{l=1}^L \sum_{w \in W} \sum_{v \in R_w}|h_v^l|
$$
where $n_f$ is the number of epochs to convergence, $R_w$ the set of vertices that are remote neighbors of vertices in $w$, $h_v^l$ is the feature vector for vertex $v$ at layer $l$.

In mini-batch training, we model a scenario where the input features cannot fit in the local memory of one worker and are partitioned among workers.
At each iteration, each worker must gather the input features of all the vertices at the bottom layer of its micro-batch.
The communication cost to convergence is:
$$
\Gamma_{\textrm{mb}} = n_m\sum_{i=1}^I \sum_{w \in W} \sum_{v \in (M_{i,w} \setminus P_w)}|h_v^0|
$$
where $n_m$ is the number of epochs to convergence using vanilla mini-batch training, $I$ is the number of iterations in an epoch, $M_{i,w}$ is the set of vertices in the micro-batch assigned to worker $w$ at iteration $i$, and $P_w$ is the partition of vertices assigned to worker $w$.



We calculate the ratio between $\Gamma_{\textrm{fg}}$ and $\Gamma_{\textrm{mb}}$ by considering the same hyperparameters for both approaches, which are the same we used for evaluating the time-to-accuracy in Section~\ref{sec:time_to_acc}.
We use Metis to partition the datasets~\cite{karypis1997metis} and run Neighborhood Sampling for one epoch to obtain $M^i_w$.

Figure~\ref{fig:data_transf} shows the ratio between $\Gamma_{\textrm{fg}}$ and $\Gamma_{\textrm{mb}}$ for different datasets.
The communication cost of full-graph training is higher than that of mini-batch training in most cases.
The gap mostly depends on the number of boundary nodes and, for mini-batch training, on the size of the last layer of the micro-batches. 
In practice, mini-batch training does not need to partition the dataset since it can replicate commonly accessed input features in the local memory of multiple workers.
In our experimental evaluation of Section~\ref{sec:perf_eval}, all datasets except orkut and papers100M are fully replicated, but partial replication is often almost as effective~\cite{ugache}.

In general, larger graphs can have a larger edge cut when they are partitioned.
This results in larger $R_w$, increasing the communication cost of full-graph training $\Gamma_{\textrm{fg}}$.
In contrast, the communication cost of mini-batch training, $\Gamma_{\textrm{mb}}$, grows with $(M_{i,w} \setminus P_w)$, which is upper bounded by the size of the sampled micro-batches, not by the size of the graph.
A similar effect is observed by increasing the number of partitions with fixed datasets, as shown in Figure~\ref{fig:data_transf}.
This increases the relative cost of $\Gamma_{\textrm{fg}}$ over $\Gamma_{\textrm{mb}}$.
\begin{figure}[t]
    \includegraphics[width=0.95 \linewidth]{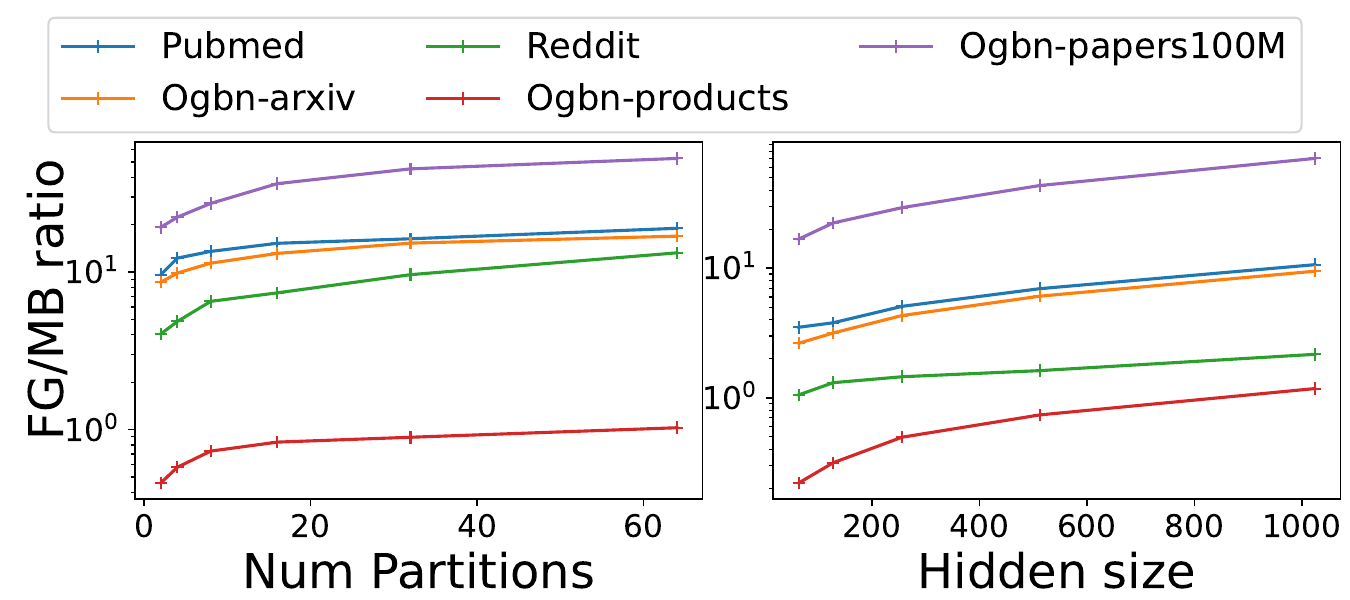}
    \caption{FG/MB ratio of communication cost. (Left: Hyperparameters in the appendix in table~\ref{tab:Communication_cost_architectures}. Right: Default 4 partitions)}
    \label{fig:data_transf}
\end{figure}

\begin{figure}[t]
    \includegraphics[width=1\linewidth]{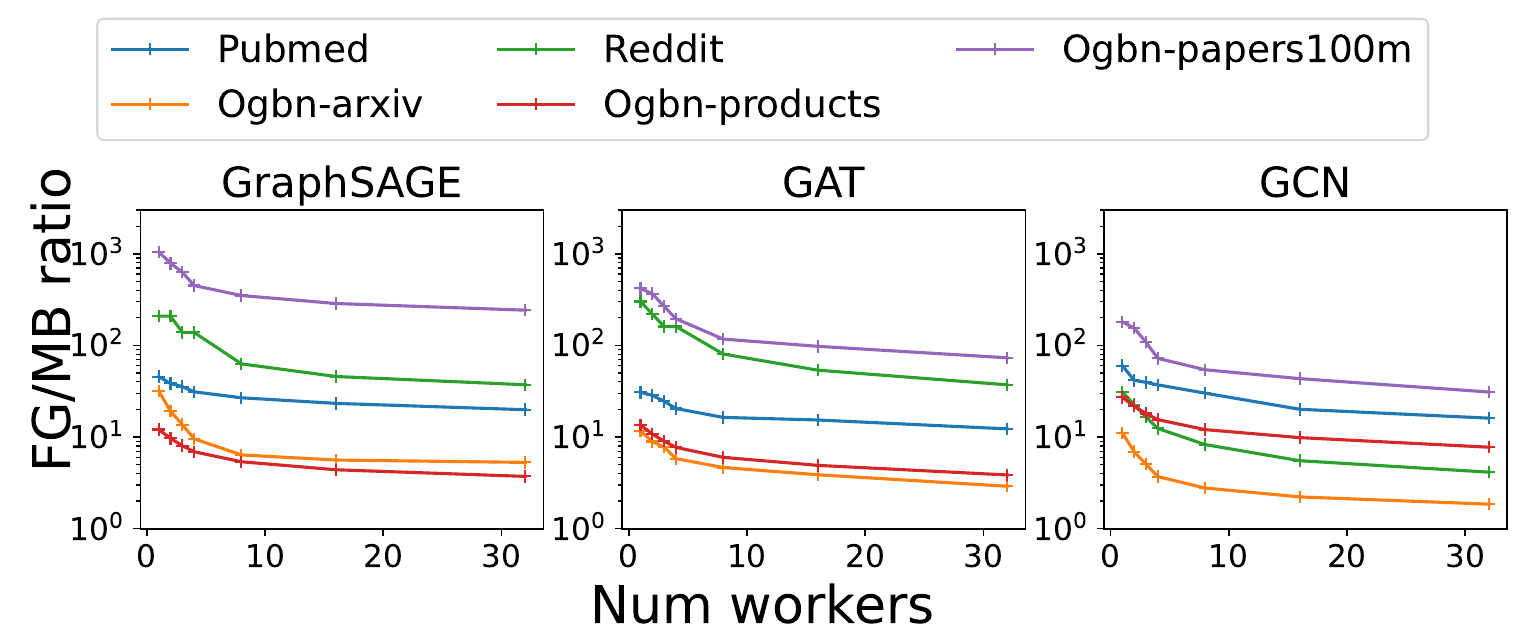}
    \caption{FG/MB ratio of computation cost for GraphSage.}
    \label{fig:flops}
\end{figure}

\spara{Computation cost.} We now analyze the computational cost of training with vanilla full-graph and mini-batch training.
We consider the same model used for the communication cost analysis.

A GNN training layer $l$ computes the features of vertices at layer $l$ based on the features of their neighbors at layer $l-1$ (see section~\ref{sec:gnn-background}).
The computational cost to the convergence of full-graph training can be expressed as: 
$$
\Theta_{\textrm{fg}} = n_f  \sum_{l=1}^L (|E^l| \cdot c_e + |V^l| \cdot c_v) (1 + \eta)
$$
where the summation expresses the cost of the forward pass: $c_e$ is the cost of generating and aggregating a message sent over one edge, $c_v$ is the cost of computing a new representation of a vertex, $V^l$ is the set of all vertices processed at layer $l$ in the epoch, and $E^l$ is the set of all edges between $V^{l-1}$ and $V^l$ processed in the epoch.
The factor $(1+\eta)$ accounts for the cost of the backward pass.

The values of $c_e$ and $c_v$ are independent of the choice of full-graph or mini-batch training. 
They depend on the GNN model (e.g. GraphSage or GAT) and its hyperparameters and they can be calculated analytically in terms of FLOPS.

The computational cost for mini-batch training is similar:
$$
\Theta_{\textrm{mb}} = n_m  \sum_{i=1}^I\sum_{w \in W}\sum_{l=1}^L (|E^l(M_{i,w})| \cdot c_e + |V^l(M_{i,w})| \cdot c_v)  (1 + \eta)
$$
where $E^l(M_{i,w})$ and $V^l(M_{i,w})$ are the number of edges and vertices at layer $l$ in the micro-batch $M_{i,w}$.

We calculate the ratio of $\Theta_{\textrm{fg}}$ and $\Theta_{\textrm{mb}}$ using the same hyperparameters as in the communication cost analysis and show it in Figure~\ref{fig:flops}.
The results reveal a large difference in terms of computational cost between mini-batch and full-graph training, especially with the largest graph, papers100M. 
In full-graph training, $\Theta_{\textrm{fg}}$ depends on $|E^l|$ and $|V^l|$, which grow as the size of the graph grows.
Mini-batch training performs message passing only on a sample of the graph, unlike full-graph training, reducing computation cost substantially.
The computation cost $\Theta_{\textrm{mb}}$ is upper-bounded by the size of $M_{i,w}$, which depends on the number of layers in the GNN and the sampling algorithm rather than on the size of the graph.

As the number of workers grows, the gap between full-graph and mini-batch training decreases in Figure~\ref{fig:flops}.
This is because the same vertex can appear in multiple micro-batches, so its hidden features are computed multiple times in an epoch.
This redundant computation does not happen with full-graph training.

\spara{Cost of sampling in mini-batch training.} 
Mini-batch training requires two pre-processing steps: sampling and data loading.
Data loading costs are included in the communication costs.
Modeling sampling costs in terms of FLOPS, as we did for the training, costs $\Theta_{\textrm{mb}}$ is challenging because sampling is a sparse and irregular computation, not a dense computation like matrix multiplication.
Therefore, to put the previous results in context, it is necessary to consider the relative computational cost of sampling over training in mini-batch training.
We measure it experimentally using DGL and Neighborhood Sampling, running both sampling and training in CPU memory to factor out the overhead of transferring data to the GPUs.
The results are shown in Figure~\ref{fig:sampling_cost}.
The computational cost of sampling is noticeable but is still lower than the training cost in most cases.
It decreases when we increase the number of workers/partitions because each epoch requires fewer iterations.

\spara{Takeaway.}
This analysis shows that vanilla full-graph training has a higher communication cost than vanilla mini-batch training when the GNN model is not very deep.
In practice, mini-batch training can replicate input features across multiple workers to eliminate or reduce communication costs.
The high communication cost of full-graph training has been the focus of prior work on full-graph training systems, as discussed in Section~\ref{sec:mb-vs-fg-optimizations}.

The analysis shows that using sampling in mini-batch training also results in a substantial reduction in computation cost compared to full-graph training systems.

\begin{figure}[t]
    \includegraphics[width=0.9 \columnwidth]{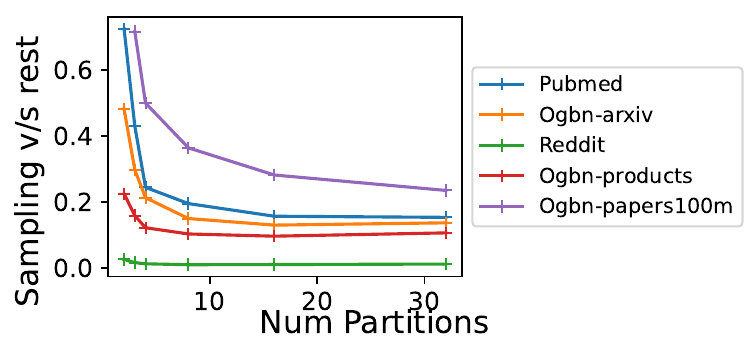}
    \caption{Fraction of epoch time spent performing sampling in mini-batch training for GraphSage.}
    \label{fig:sampling_cost}
   \vspace{-0.5cm}
\end{figure}

}
\section{Conclusion}
\label{sec:conclusion}
\finalrevision{Our work has evaluated various full-graph and mini-batch systems, encompassing a wide range of optimizations (see Figure~\ref{fig:baselines}).}
\revision{
A common rationale for using full-graph instead of mini-batch training is to achieve higher accuracy by avoiding the information loss resulting from sampling.
In our evaluation, we did not observe these expected benefits.
Mini-batch training systems not only achieve a lower time-to-accuracy across all models, datasets, and hardware configurations we consider but also a comparable or even higher accuracy with proper hyperparameter tuning.

\finalrevision{Among the optimizations we considered for mini-batch training, GPU-based sampling and GPU caching should be used whenever feasible. The choice and configuration of the sampling algorithm is critical for performance and accuracy.}
The common Neighborhood Sampling algorithm is able to consistently achieve good accuracy, but it still shows a slightly larger variance between training runs than vanilla full-graph training.
Other sampling algorithms we evaluated are not always able to converge to a high accuracy, depending on the dataset.
Distributed sampling over multiple hosts can also become a bottleneck, especially when the network is slow.
Systems that allow ML practitioners to implement new sampling algorithms and run them on GPUs, such as~\cite{nextdoor,c-saw,gsampler,wang2021skywalker}, can support future innovations in efficient sampling algorithms. 


\finalrevision{For full-graph training systems, the algorithmic optimizations we considered can substantially improve performance with minimal or no impact on accuracy.}
Our performance evaluation and cost model analysis reveal that in addition to the well-known communication bottleneck addressed by much previous work, future optimizations should also reduce the computation cost gap with mini-batch training.
A natural open question is whether combining the full-batch training pipeline of Figure~\ref{fig:fg_mb_training_diagram}(a) with more aggressive sampling, forgoing full-graph aggregation, can be more effective than mini-batch training.
In our hyperparameter tuning, however, we found that the accuracy of mini-batching typically peaks with batch sizes around 1024/4096 and does not grow further with larger values, so the advantage of using full-batch training is unclear.
}




\bibliographystyle{ACM-Reference-Format}
\bibliography{ref}
\newpage
\appendix
\onecolumn
\section*{Appendix}
\label{sec:appendix}
The results for the highest achievable accuracy for the GAT and GCN model are shown in Table~\ref{tab:convergence_gat} and Table~\ref{tab:convergence_gcn}. 
Given a dataset, we must compare the accuracies obtained for two combinations: (\textsc{FG-search}, \textsc{FG-train}) for full-graph training and (\textsc{MB-search}, \textsc{MB-train}) for mini-batch training. The results are averaged after 5 runs.

Table~\ref{parameter_search_space} shows the search space of our hyperparameter search.
Table~\ref{tab:best_architectures} shows the best architectures obtained for every model-dataset combination.
Table~\ref{tab:TTA_architectures} shows the architectures used for TTA comparisons.
Tables~\ref{tab:scalability_architectures_single_host} and ~\ref{tab:scalability_architectures_multi_host} show the architectures used for scalability comparisons.
Table~\ref{tab:Communication_cost_architectures} shows the architectures used for communication cost experiments.

Figure~\ref{fig:GCN_robustness} and  Figure~\ref{fig:GAT_robustness} illustrate the best test accuracy when training the GCN and GAT respectively.
The top row shows mini-batch training experiments and the bottom row shows synchronous full-graph training.

\begin{table}[b]
\caption{Convergence for GAT model}
\label{tab:convergence_gat}
\vskip -0.15in
    \begin{center}
    \begin{small}
\begin{tabular}{ | m{5em}| m{3.5em}| m{3.5em}| m{3.5em} |m{3.5em} |m{3.5em} |m{3.5em} |m{3.5em}| m{3.5em}|} 
\toprule
 & \multicolumn{2}{c|}{Pubmed}  & \multicolumn{2}{c|}{Ogbn-arxiv} & \multicolumn{2}{c|}{Reddit} & \multicolumn{2}{c|}{Ogbn-products}\\
\midrule
 & FG-search & MB-search & FG-search & MB-search & FG-search & MB-search & FG-search & MB-search \\
\midrule
 FG-train & $77.70 \pm 00.50$   & $77.20 \pm 00.34$ & $71.74 \pm 00.22 $ & $67.77 \pm 00.31$ & $95.17 \pm 00.20$ & $93.00 \pm 00.23$ & $ 74.43 \pm 00.05$ & $73.33 \pm 00.15$ \\

MB-train & $75.80 \pm 00.06$ & $78.50 \pm 00.03$ & $70.19 \pm 00.16$ & $70.63 \pm 00.12$ & $ 94.58 \pm 00.14$ & $95.58 \pm 00.13$ & $74.27 \pm 00.07$ & $74.57 \pm 00.07$ \\
\bottomrule
\end{tabular}
    \end{small}
    \end{center}
    \vskip -0.1in
\end{table}

\begin{table}[b]
\caption{Convergence for GCN model}
\label{tab:convergence_gcn}
\vskip -0.15in
    \begin{center}
    \begin{small}
    \begin{sc}    
\begin{tabular}{ |m{5em}| m{3.5em}| m{3.5em}| m{3.5em} |m{3.5em}| m{3.5em} |m{3.5em}| m{3.5em} |m{3.5em}|} 
\toprule
 & \multicolumn{2}{c|}{Pubmed}  & \multicolumn{2}{c|}{Ogbn-arxiv} & \multicolumn{2}{c|}{Reddit} & \multicolumn{2}{c|}{Ogbn-products}\\
\midrule
& FG-search & MB-search & FG-search & MB-search & FG-search & MB-search & FG-search & MB-search \\
\midrule
 FG-train & $79.40 \pm 00.12$   & $75.10 \pm 00.37$ & $71.54 \pm 00.21 $ & $69.32 \pm 00.22$ & $94.67 \pm 00.20$ & $93.64 \pm 00.10$ & $75.71 \pm 00.18$ & $73.81 \pm 00.13$ \\

MB-train & $78.20 \pm 00.16$ & $78.70 \pm 00.23$ & $70.54 \pm 00.16$ & $71.00 \pm 00.16$ & $93.31 \pm 00.14$ & $94.41 \pm 00.14$ & $75.27 \pm 00.07$ & $79.31 \pm 00.04$ \\
\bottomrule
\end{tabular}
    \end{sc}
    \end{small}
    \end{center}
    \vskip -0.1in
\end{table}
\begin{table}[t]
\caption{Hyperparameter search space}
\label{parameter_search_space}
    \begin{center}
    \begin{small}
    \begin{tabular}{|l|r|}
    \toprule
    Parameter & range/values \\
    \midrule
    Number of Hidden Layers & [2, 10]  \\
    Size of hidden layers   & [16, 1024] \\
    Aggregator (GraphSage only) & \{ mean, gcn, pool \} \\
    Number of heads (GAT only) & [1, 12] \\
    \midrule
    Sampling Fanout (MB only) & [4, 50] \\
    \midrule
    Learning rate           & [0.0001, 0.01] \\
    Dropout                 & [0.2, 0.9] \\
    Micro-batch size (MB only)            & [32, 4096] \\
    \bottomrule
    \end{tabular}
    \end{small}
    \end{center}
\end{table}
\begin{table*}[b]
\caption{Best architectures}
\label{tab:best_architectures}
\vskip -0.15in
    \begin{center}
    \begin{small}
    \begin{sc}    
\begin{tabular}{|*{7}{c|}}
\toprule
 Model & Training technique & Architecture & Pubmed & Ogbn-arxiv & Reddit &  Ogbn-products \\
\midrule
\multirow{4}{*}{GraphSAGE} & \multirow{2}{*}{FG} & Layers & 4  & 3 & 4 & 5 \\
& & Hidden size & 64  & 512 & 1024 & 512 \\
& \multirow{2}{*}{MB} & Layers & 2 & 5  & 4 & 5 \\
& & Hidden size &  256 & 128 & 512 & 256 \\
& & Fanout & 10 & 20 & 5 & 5 \\
\midrule
\multirow{6}{*}{GAT} & \multirow{3}{*}{FG} & Layers & 3  & 3 & 2 &  2\\
& & Hidden size & 1024  & 1024 & 1024 & 256 \\
& & Num heads & 1 & 2 & 2 & 2 \\
& \multirow{3}{*}{MB} & Layers & 2 &  4 & 2 &  3\\
& & Hidden size & 1024  & 256 & 512 & 128\\
& & Num heads & 4 & 2 & 2 & 2 \\
& & Fanout & 10 & 10 & 10 & 5 \\
\midrule
\multirow{4}{*}{GCN} & \multirow{2}{*}{FG} & Layers & 2  & 2 & 2 & 3 \\
& & Hidden size & 512  & 1024 & 1024 & 512 \\
& \multirow{2}{*}{MB} & Layers &6 & 2  & 2 & 2 \\
& & Hidden size &  64 & 1024 & 512 & 512 \\
& & Fanout & 10 & 15 & 15 & 5 \\

\bottomrule
\end{tabular}
    \end{sc}
    \end{small}
    \end{center}
    \vskip -0.1in
\end{table*}

\begin{table*}[b]
\caption{Architectures used for TTA experiments (same for FG and MB).}
\label{tab:TTA_architectures}
\vskip -0.15in
    \begin{center}
    \begin{small}
    \begin{sc}    
\begin{tabular}{|*{7}{c|}}
\toprule
 Model & Architecture & Pubmed & Ogbn-arxiv & Reddit &  Ogbn-products & Ogbn-papers100M \\
\midrule
\multirow{3}{*}{GraphSAGE} & Layers & 3  & 2 & 4 & 5 &  2 \\
&  Hidden size & 256  & 512 & 1024 & 256 & 128 \\
&  Fanout & 10 & 20 & 5 & 5 & 5 \\
\midrule
\multirow{4}{*}{GAT} & Layers & 3  & 3 & 2 & 3 & - \\
&  Hidden size & 256  & 1024 & 1024 & 128 &  - \\
&  Fanout & 10  & 10 & 10 & 5 &  - \\
&  Num Heads & 4 & 6 & 4 & 3 &  - \\
\bottomrule
\end{tabular}
    \end{sc}
    \end{small}
    \end{center}
    \vskip -0.1in
\end{table*}

\begin{table*}[b]
\caption{Architectures used for Scalability experiments single host (same for FG and MB).}
\label{tab:scalability_architectures_single_host}
\vskip -0.15in
    \begin{center}
    \begin{small}
    \begin{sc}    
\begin{tabular}{|*{8}{c|}}
\toprule
 Model & Architecture  & Ogbn-arxiv & Reddit &  Ogbn-products & Orkut \\
\midrule
\multirow{4}{*}{GAT} & Layers  & 3 & 2 & 3 & 3 \\
&  Hidden size & 1024 & 1024 & 128 & 128 \\
&  Num Heads & 6 & 4 & 3 & 2 \\
&  Fanout  & 25 & 20 & 20 & 5 \\
\bottomrule
\end{tabular}
    \end{sc}
    \end{small}
    \end{center}
    \vskip -0.1in
\end{table*}

\begin{table*}[b]
\caption{Architectures used for Scalability experiments for 3 hosts (same for FG and MB).}
\label{tab:scalability_architectures_multi_host}
\vskip -0.15in
    \begin{center}
    \begin{small}
    \begin{sc}    
\begin{tabular}{|*{8}{c|}}
\toprule
 Model & Architecture  & Orkut & Ogbn-papers100m \\
\midrule
\multirow{3}{*}{GraphSAGE} & Layers & 3  & 2 \\
&  Hidden size & 512  & 512  \\
&  Fanout & 10 & 10  \\
\bottomrule
\end{tabular}
    \end{sc}
    \end{small}
    \end{center}
    \vskip -0.1in
\end{table*}

\begin{table*}[b]
\caption{Architectures used for Communication cost experiments (Same for FG and MB).}
\label{tab:Communication_cost_architectures}
\vskip -0.15in
    \begin{center}
    \begin{small}
    \begin{sc}    
\begin{tabular}{|*{7}{c|}}
\toprule
 Model & Architecture & Pubmed & Ogbn-arxiv & Reddit &  Ogbn-products & Ogbn-papers100M \\
\midrule
\multirow{3}{*}{GraphSAGE} & Layers & 3  & 2 & 4 & 5 &  2 \\
&  Hidden size & 256  & 512 & 1024 & 256 & 128 \\
&  Fanout & 10 & 20 & 5 & 5 & 5 \\
\bottomrule
\end{tabular}
    \end{sc}
    \end{small}
    \end{center}
    \vskip -0.1in
\end{table*}

\begin{table*}[t]
\caption{Time to accuracy for GAT on a single NVLink host with 4 GPUs. Architectures (\#Layers, hidden size, \#heads, fanout): ogbn-arxiv - (3, 1024, 6, 25), Reddit - (2, 1024, 4, 20), ogbn-products - (3, 128, 3, 20). ET - Epoch time, TTA - Time to accuracy. For ogbn-papers100m DGL uses UVA mode for sampling and Quiver performs distributed caching.}
\label{tab:time_to_acc_nvlink_gat}
    \begin{center}
    \begin{small}
    \begin{sc}    

\begin{tabular}{ |l|l*{3}{|r}|} 
\toprule
 & System & ogbn-arxiv & reddit & ogbn-products  \\

\midrule
\multirow{3}{7em}{ET (s)} 
& PipeGCN (FG, Sync)   & 0.625 & 1.801 & 1.033 \\
& PipeGCN (FG, Async)  & 0.469 & 1.312 & 0.614 \\
& BNS-GCN (FG, Sync)   & 0.420 & 1.395 & 0.629 \\
& DGL (MB, Sync)       & 3.871 & 6.510 & 2.730 \\
& Quiver (MB, Sync)    & 3.596 & 6.779 & 2.974 \\
\midrule
\multirow{3}{7em}{TTA (s)} 
& PipeGCN (FG, Sync)  & 275.04 & 1728.58 & 413.28  \\
& PipeGCN (FG, Async) & 211.14 & 1312.15 & 233.24  \\
& BNS-GCN (FG, Sync)  & 151.20 & 1395.40 & 132.13 \\
& DGL (MB, Sync)      & 42.58 & 97.65 & 40.95 \\
& Quiver (MB, Sync)   & 35.96 & 101.68 & 44.61 \\
      \bottomrule
\end{tabular}
    \end{sc}
    \end{small}
    \end{center}
    \vskip -0.1in
\end{table*}

Table~\ref{tab:time_to_acc_nvlink_gat} shows the results of the time-to-accuracy experiment for the GAT model on a single host with NVLink connectivity.  
Table~\ref{tab:accuracies_gat_all} and ~\ref{tab:accuracies_gcn_all} show the accuracies achieved for GAT and GCN models for different training techniques an optimizations.

Table~\ref{tab:scalability_gat_m40_single_host} shows the results of the scalability experiments for the GAT model on PCIe setup and 
table~\ref{tab:scalability_graphsage_a100_single_host} shows the results of the scalability experiment on the NVLink setup. Table~\ref{tab:time_to_acc_papers} shows the results of the time-to-accuracy experiments for the GraphSAGE model on 3 PCIe hosts.

\begin{figure}
    \centering
    \includegraphics[width=\textwidth]{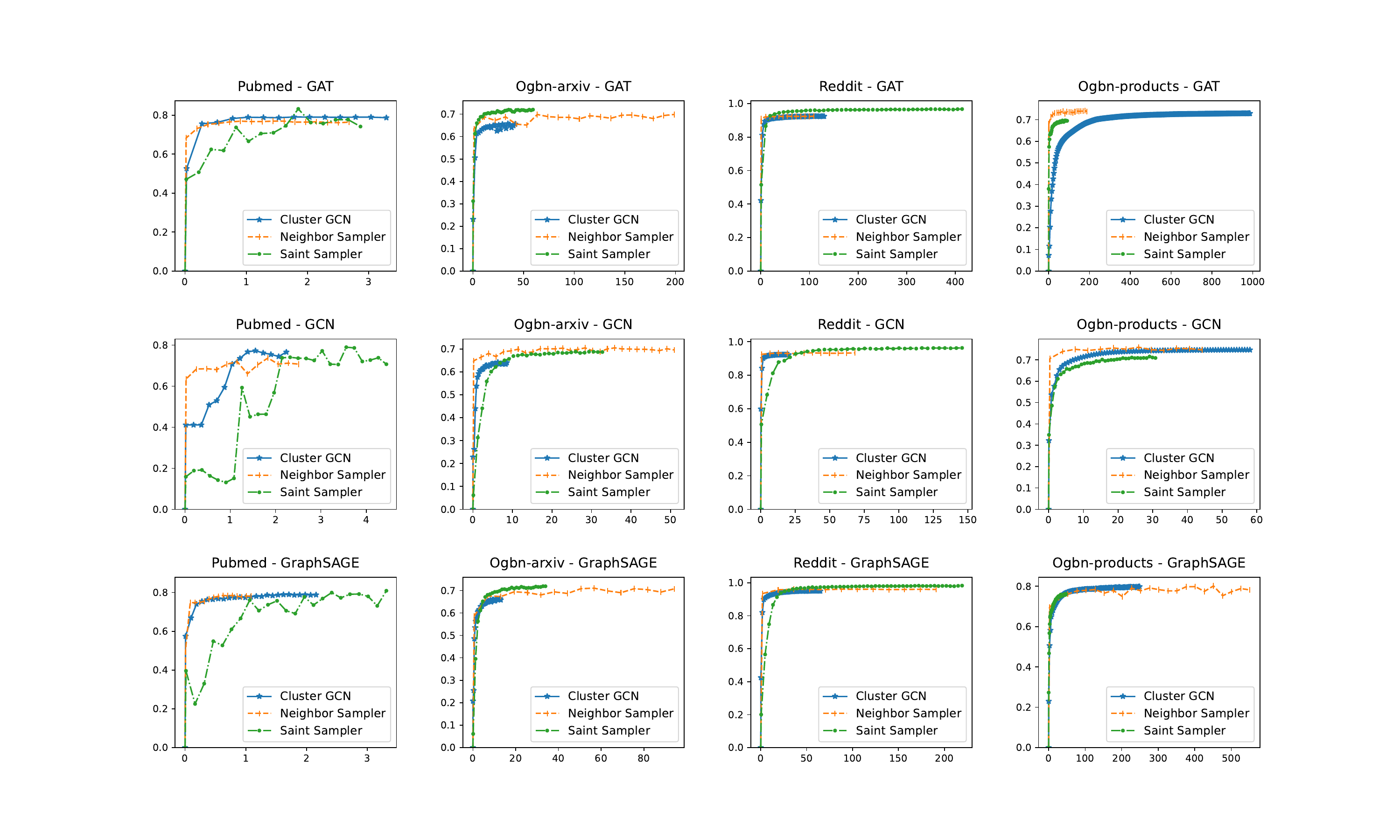}
    \caption{Test accuracy achieved for the best architecture for GraphSAGE, GAT and GCN models with NeighborSampler, ClusterGCN and SAINT Sampler.}
    \label{fig:best_arc_samplers}
\end{figure}

\begin{figure*}
    \centering
    \includegraphics[width=1\textwidth]{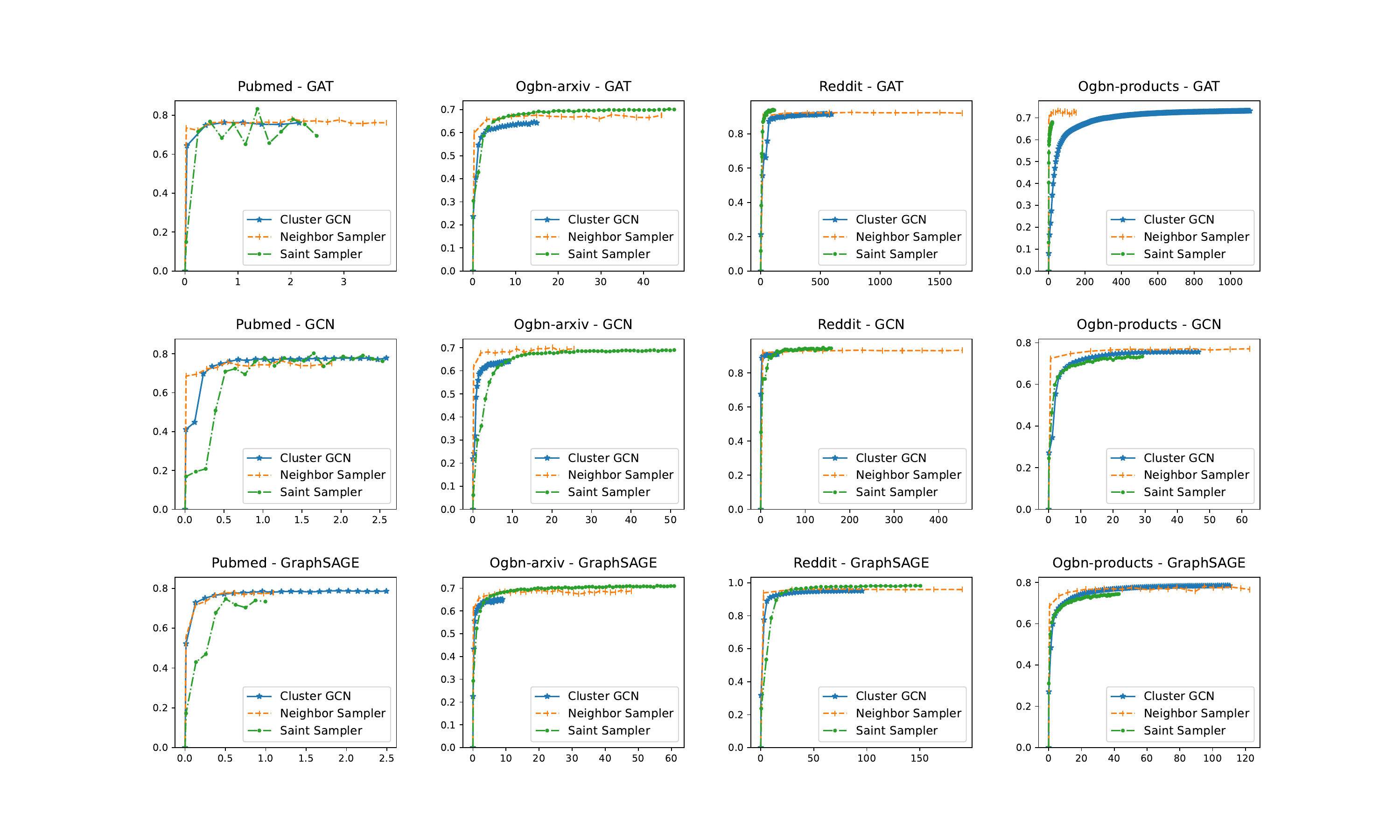}
    \caption{Time to accuracy plots for GraphSAGE, GAT and GCN models with NeighborSampler, ClusterGCN and SAINT Sampler.}
    \label{fig:tta_samplers}
\end{figure*}

Tables~\ref{tab:time_to_acc_graphsage_samplers}, \ref{tab:time_to_acc_gat_samplers} and \ref{tab:time_to_acc_gcn_samplers} show the epoch time and time to accuracy for GraphSAGE, GAT and GCN respectively for different sampling optimizations. Figure~\ref{fig:best_arc_samplers} shows the best accuracy achieved and \ref{fig:tta_samplers} shows the time-to-accuracy plots for GraphSAGE, GAT and GCN models with different mini-batch optimizations (Neighbor sampler, ClusterGCN and SAINT sampler).

\begin{table}[b]
\caption{Best accuracies (\%) for GAT model}
\label{tab:accuracies_gat_all}
\vskip -0.15in
    \begin{center}
    \begin{small}
\begin{tabular}{ |l*{5}{|r} |} 
\toprule
& Pubmed & Ogbn-arxiv & Reddit & Ogbn-products \\
\midrule
FG & $77.70 \pm 0.50 $ & $71.74 \pm 0.22 $ & $95.17 \pm 0.20 $ & $74.43 \pm 0.05 $ \\
PipeGCN & $77.60 \pm 0.37$ & $70.34 \pm 0.10$ & $93.63 \pm 0.07 $ & $74.08 \pm 0.12 $ \\
BNS-GCN & $77.70 \pm 0.10 $ & $71.75 \pm 0.70 $ & $94.23 \pm 0.02$  & $74.50 \pm 0.26 $ \\
\hline
NS & $78.50 \pm 0.03$  & $70.63 \pm 0.12$ & $95.58 \pm 0.13$ & $74.57 \pm 0.07$ \\
ClusterGCN & $80.07 \pm 0.19 $ & $63.30 \pm 0.09 $ & $92.49 \pm 0.75 $ & $73.27 \pm 0.29$\\ 
GraphSaint &$82.76 \pm 0.95$ & $72.58 \pm 0.17 $ & $96.96 \pm 0.06 $ & $70.47 \pm 0.16 $ \\ 
\bottomrule
\end{tabular}
    \end{small}
    \end{center}
    \vskip -0.1in
\end{table}

\begin{table}[b]
\caption{Best accuracies (\%) for GCN model}
\label{tab:accuracies_gcn_all}
\vskip -0.15in
    \begin{center}
    \begin{small}
\begin{tabular}{ |l*{5}{|r} |} 
\toprule
& Pubmed & Ogbn-arxiv & Reddit & Ogbn-products \\
\midrule
FG & $79.40 \pm 0.12$& $71.54 \pm 0.21 $& $94.67 \pm 0.20 $& $75.71 \pm 0.18 $\\
PipeGCN & $77.00 \pm 0.81$& $69.64 \pm 0.27 $& $93.91 \pm 0.10 $& $74.25 \pm 0.34 $\\
BNS-GCN &$77.70 \pm 0.71 $& $70.75 \pm 0.36$ & $93.23 \pm 0.09 $& $74.50 \pm 0.32 $\\
\hline
NS & $78.70 \pm 0.23$ & $71.00 \pm 0.16 $& $94.41 \pm 0.14$& $79.31 \pm 0.04 $\\
Cluster & $77.50 \pm 1.98 $ & $63.24 \pm 0.10 $ & $92.42 \pm 0.03$ & $73.71 \pm 0.08$\\ 
GraphSaint &  $81.70 \pm 0.71$ & $69.72 \pm 0.08 $ &$96.88 \pm 0.06$ & $72.65 \pm 0.13 $ \\ 
\bottomrule
\end{tabular}
    \end{small}
    \end{center}
    \vskip -0.1in
\end{table}

\begin{table*}[t]
\caption{Single-host scalability with the GAT model using the PCIe host. Times in seconds. (\#Layers, hidden size, \#heads, fanout): ogbn-arxiv - (3, 1024, 2, 25), Reddit - (2, 1024, 1, 15), ogbn-products - (3, 128, 1, 10), Orkut - (2, 128, 1, 10). ET - Epoch time, TTA - Time To Accuracy.}
\label{tab:scalability_gat_m40_single_host}
    \begin{center}
    \begin{small}
    \begin{sc}    
\begin{tabular}{|m{9em} | m{6em}| >{\raggedleft\arraybackslash}m{3em} >{\raggedleft\arraybackslash}m{3em} >{\raggedleft\arraybackslash}m{3em} | >{\raggedleft\arraybackslash}m{3em} >{\raggedleft\arraybackslash}m{3em} >{\raggedleft\arraybackslash}m{3em}| >{\raggedleft\arraybackslash}m{3em} >{\raggedleft\arraybackslash}m{3em} >{\raggedleft\arraybackslash}m{3em}| >{\raggedleft\arraybackslash}m{3em}|}
\toprule
Dataset &  & \multicolumn{3}{c|}{ogbn-arxiv} & \multicolumn{3}{c|}{Reddit} & \multicolumn{3}{c|}{ogbn-products} & Orkut \\
\midrule 
System & \# of GPUs & ET & TTA & TTA Speed-up & ET & TTA  & TTA Speed-up & ET & TTA  & TTA Speed-up & ET  \\
\midrule 
\multirow{4}{*}{Full-Graph} 
& 1 & 2.225 & 979.09 & 1.0x & 5.857 & 5622.53 & 1.0x & 3.510 & 1404.12 & 1.0x & OOM  \\
& 2 & 1.514 & 665.98 & 1.5x & 4.035 & 3873.98 & 1.5x & 3.381 & 1352.32 & 1.0x & 2.726  \\
& 3 & 1.323 & 581.98 & 1.7x & 4.034 & 3872.83 & 1.5x & 2.748 & 1099.16 & 1.3x & 2.425   \\
& 4 & 1.072 & 471.55 & 2.1x & 2.890 & 2774.78 & 2.0x & 2.452 & 980.60 & 1.4x & 2.323   \\
\midrule
\multirow{4}{*}{PipeGCN} 
& 1 & 2.218 & 998.01 & 1.0x & 5.856 & 5856.10 & 1.0x & 3.503 & 1331.10 & 1.0x & OOM   \\
& 2 & 1.334 & 600.08 & 1.7x & 3.269 & 3269.20 & 1.8x & 2.762 & 1049.37 & 1.3x & 2.479  \\
& 3 & 1.026 & 461.79 & 2.2x & 3.109 & 3109.20 & 1.9x & 1.972 & 749.17 & 1.8x & 1.898  \\
& 4 & 0.805 & 362.03 & 2.8x & 1.879 & 1879.10 & 3.1x & 1.523 & 578.78 & 2.3x & 1.499  \\
\midrule
\multirow{4}{*}{BNS-GCN} 
& 1 & 2.337 & 841.43 & 1.0x & 6.232 & 6231.60 & 1.0x & 4.010 & 842.10 & 1.0x & OOM  \\
& 2 & 1.345 & 484.02 & 1.7x & 3.472 & 3472.20 & 1.8x & 3.466 & 727.82 & 1.2x & 2.860   \\
& 3 & 0.980 & 352.76 & 2.4x & 2.697 & 2697.10 & 2.3x & 2.366 & 496.92 & 1.7x & 2.278  \\
& 4 & 0.778 & 280.08 & 3.0x & 2.193 & 2192.70 & 2.8x & 1.710 & 359.06 & 2.3x & 1.583  \\
\midrule
\multirow{4}{*}{DGL} 
& 1 & 15.401 & 154.01 & 1.0x & 17.468 & 262.01 & 1.0x & 18.394 & 367.88 & 1.0x & 22.465 \\
& 2 & 8.888 & 88.88 & 1.7x & 8.575 & 128.62 & 2.0x & 9.328 & 186.57 & 2.0x & 11.580\\
& 3 & 6.146 & 61.46 & 2.5x & 5.722 & 85.83 & 3.1x & 6.246 & 131.17 & 2.8x & 7.544  \\
& 4 & 4.627 & 46.27 & 3.3x & 4.594 & 68.91 & 3.8x & 4.715 & 99.02 & 3.7x & 6.444  \\
\midrule\
\multirow{4}{*}{Quiver} 
& 1 & 16.230 & 162.30 & 1.0x & 18.914 & 283.71 & 1.0x & 17.457 & 349.15 & 1.0x & 24.222  \\
& 2 & 8.286 & 91.15 & 1.8x & 9.002 & 135.03 & 2.1x & 8.783 & 175.66 & 2.0x & 14.134 \\
& 3 & 5.737 & 68.85 & 2.4x & 6.972 & 111.55 & 2.5x & 5.633 & 118.29 & 3.0x & 9.048  \\
& 4 & 4.256 & 51.08 & 3.2x & 4.595 & 87.31 & 3.2x & 4.695 & 98.59 & 3.5x & 6.695  \\
\bottomrule
\end{tabular}
    \end{sc}
    \end{small}
    \end{center}
    \vskip -0.1in
\end{table*}

\begin{table*}[t]
\caption{Single-host scalability with the GraphSAGE model using the NVLink host. Times in seconds. (\#Layers, hidden size, fanout): ogbn-arxiv - (3, 1024, 25), Reddit - (2, 1024, 20), ogbn-products - (3, 128, 20), Orkut - (3, 128, 5). ET - Epoch time, TTA - Time to Accuracy.}
\label{tab:scalability_graphsage_a100_single_host}
    \begin{center}
    \begin{small}
    \begin{sc}    
\begin{tabular}{|m{9em} | m{6em}| >{\raggedleft\arraybackslash}m{3em} >{\raggedleft\arraybackslash}m{3em} >{\raggedleft\arraybackslash}m{3em} | >{\raggedleft\arraybackslash}m{3em} >{\raggedleft\arraybackslash}m{3em} >{\raggedleft\arraybackslash}m{3em}| >{\raggedleft\arraybackslash}m{3em} >{\raggedleft\arraybackslash}m{3em} >{\raggedleft\arraybackslash}m{3em}| >{\raggedleft\arraybackslash}m{3em} |}
\toprule
Dataset &  & \multicolumn{3}{c|}{ogbn-arxiv} & \multicolumn{3}{c|}{Reddit} & \multicolumn{3}{c|}{ogbn-products} & Orkut \\
\midrule 
System & \# of GPUs & ET & TTA & TTA Speed-up & ET & TTA  & TTA Speed-up & ET & TTA  & TTA Speed-up & ET \\
\midrule 
\multirow{4}{*}{PipeGCN (Sync)} 
 & 1 & 0.181 & 39.90 & 1.0x & 0.433 & 121.16 & 1.0x & 0.270 & 107.90 & 1.0x & 0.466\\
 & 2 & 0.182 & 40.04 & 1.0x & 0.746 & 208.78 & 0.6x & 0.398 & 159.28 & 0.7x & 0.452\\
 & 3 & 0.315 & 69.24 & 0.6x & 0.735 & 205.68 & 0.6x & 0.306 & 122.34 & 0.9x & 1.312\\
 & 4 & 0.211 & 46.51 & 0.9x & 0.776 & 217.40 & 0.6x & 0.518 & 207.20 & 0.5x & 0.995\\
\midrule
\multirow{4}{*}{PipeGCN (Async)} 
 & 1 & 0.182 & 56.28 & 1.0x & 0.432 & 120.94 & 1.0x & 0.270 & 102.48 & 1.0x & 0.467\\
 & 2 & 0.098 & 30.38 & 1.9x & 0.255 & 71.41 & 1.7x & 0.164 & 62.24 & 1.6x & 0.281\\
 & 3 & 0.071 & 21.97 & 2.6x & 0.212 & 59.37 & 2.0x & 0.114 & 43.17 & 2.4x & 0.234\\
 & 4 & 0.057 & 17.53 & 3.2x & 0.130 & 36.42 & 3.3x & 0.089 & 33.71 & 3.0x & 0.183\\
\midrule
\multirow{4}{*}{BNS-GCN (Sync)} 
 & 1 & 0.188 & 50.74 & 1.0x & 0.369 & 58.97 & 1.0x & 0.363 & 76.31 & 1.0x & 0.536\\
 & 2 & 0.106 & 28.71 & 1.8x & 0.227 & 36.29 & 1.6x & 0.219 & 46.03 & 1.7x & 0.322\\
 & 3 & 0.083 & 22.45 & 2.3x & 0.183 & 29.33 & 2.0x & 0.162 & 34.00 & 2.2x & 0.289\\
 & 4 & 0.069 & 18.56 & 2.7x & 0.157 & 25.07 & 2.4x & 0.136 & 28.64 & 2.7x & 0.225\\
\midrule
\multirow{4}{*}{DGL (Sync)} 
 & 1 & 1.519 & 7.60 & 1.0x & 1.571 & 7.85 & 1.0x & 2.819 & 42.28 & 1.0x & 7.182\\
 & 2 & 0.842 & 4.21 & 1.8x & 0.846 & 4.23 & 1.9x & 1.389 & 20.84 & 2.0x & 3.606\\
 & 3 & 0.577 & 2.89 & 2.6x & 0.576 & 2.88 & 2.7x & 0.926 & 13.90 & 3.0x & 2.390\\
 & 4 & 0.457 & 2.28 & 3.3x & 0.490 & 2.45 & 3.2x & 0.699 & 10.49 & 4.0x & 1.842\\
\midrule
\multirow{4}{*}{Quiver (Sync)} 
& 1 & 0.906 & 9.06 & 1.0x & 2.478 & 12.39 & 1.0x & 2.463 & 36.94 & 1.0x & 6.301\\
& 2 & 0.481 & 4.81 & 1.9x & 1.278 & 6.39 & 1.9x & 1.250 & 18.75 & 2.0x & 3.317\\
& 3 & 0.333 & 3.33 & 2.7x & 0.901 & 4.51 & 2.8x & 0.861 & 12.92 & 2.9x & 2.279\\
& 4 & 0.260 & 2.60 & 3.5x & 0.731 & 3.66 & 3.4x & 0.643 & 9.65 & 3.8x & 1.743\\
\bottomrule
\end{tabular}
    \end{sc}
    \end{small}
    \end{center}
    \vskip -0.1in
\end{table*}

\begin{table}[t]
\caption{Time to accuracy (TTA) and epoch time (ET) for GraphSage and ogbn-papers100M (3 PCIe hosts)}
\label{tab:time_to_acc_papers}
    \begin{center}
    \begin{small}
\begin{tabular}{ | m{8em} | m{9em} |>{\raggedleft\arraybackslash}m{5em}|} 
\toprule
 & System & Time (s)\\
\midrule
\multirow{3}{7em}{Epoch time (s)} & Full-Graph & 22.00 \\
& PipeGCN& \textbf{1.03} \\
& BNS-GCN&  2.46 \\
\cline{2-3}
& DistDGL & 38.37 \\
\midrule
\multirow{3}{7em}{Time to accuracy (s)} & Full-Graph &  19141.48 \\
& PipeGCN &  956.34 \\
& BNS-GCN&  859.26 \\
\cline{2-3}
& DistDGL&  \textbf{767.38} \\
      \bottomrule
\end{tabular}
    \end{small}
    \end{center}
    \vskip -0.1in
\end{table}

\section{Ease of Hyperparameter Search}


\begin{figure*}[h]
    \centering
    \includegraphics[width=\textwidth]{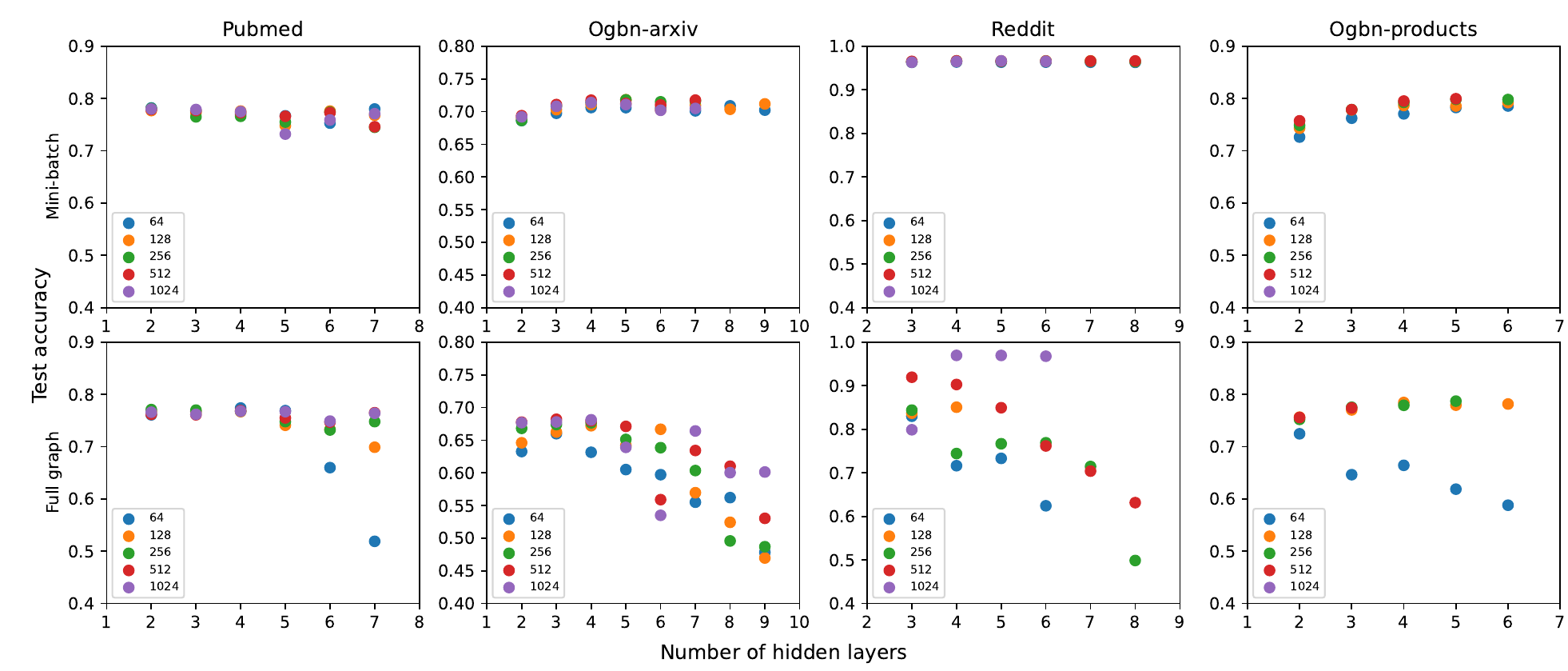}
    \vskip -0.1in
    \caption{Sensitivity to the number of layers and hidden size for GraphSAGE model.}
    \label{fig:all_robustness}
\end{figure*}

\begin{figure*}[b]
    \centering
    \includegraphics[width=\textwidth]{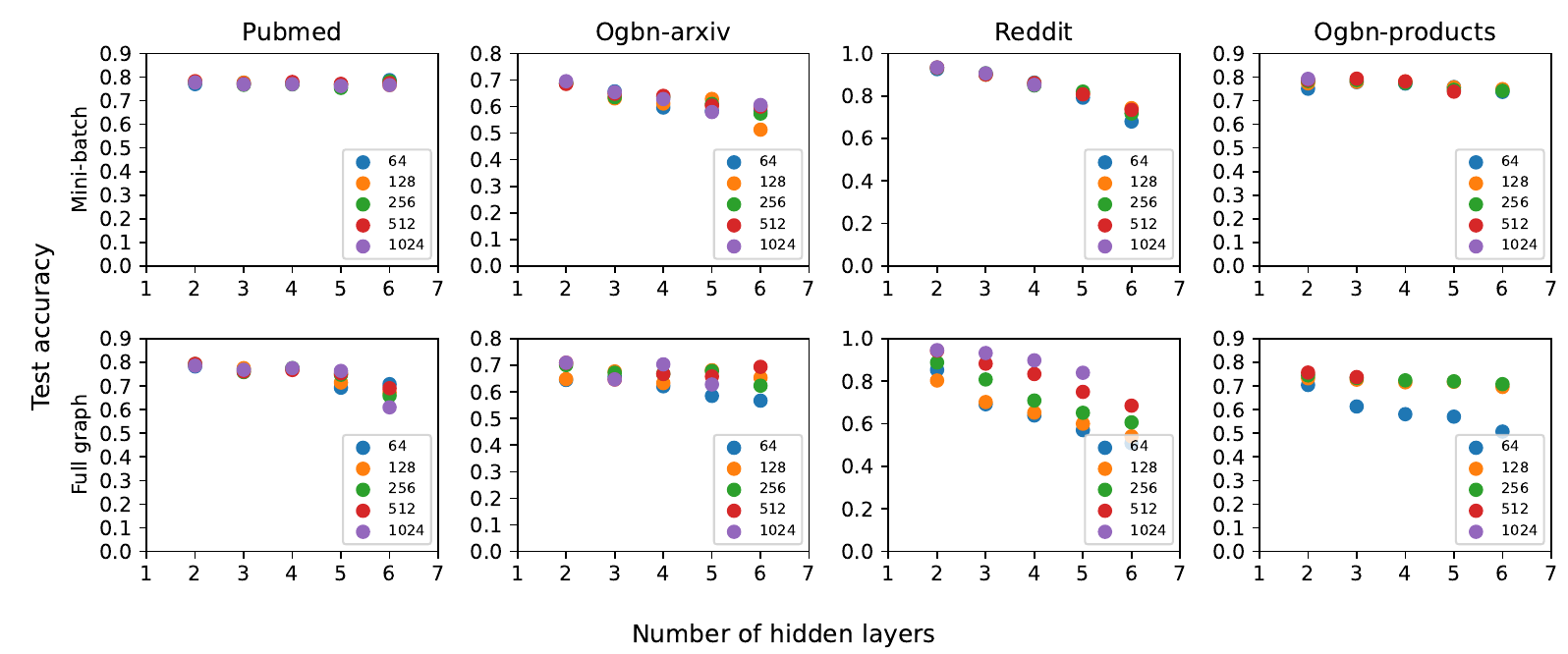}
    \vskip -0.1in
    \caption{Ablation study with varying number of layers and hidden size for GCN model. Legend represents the size of each hidden layer. The X-axis shows the number of layers. The Y-axis shows the test accuracy. The top row is for mini-batch training and the bottom row is for full graph training. We observe that mini-batch training is robust to architectural changes, whereas full-graph training has a lot of variation.}
    \label{fig:GCN_robustness}
\end{figure*}

\begin{figure*}[b]
    \centering
    \includegraphics[width=\textwidth]{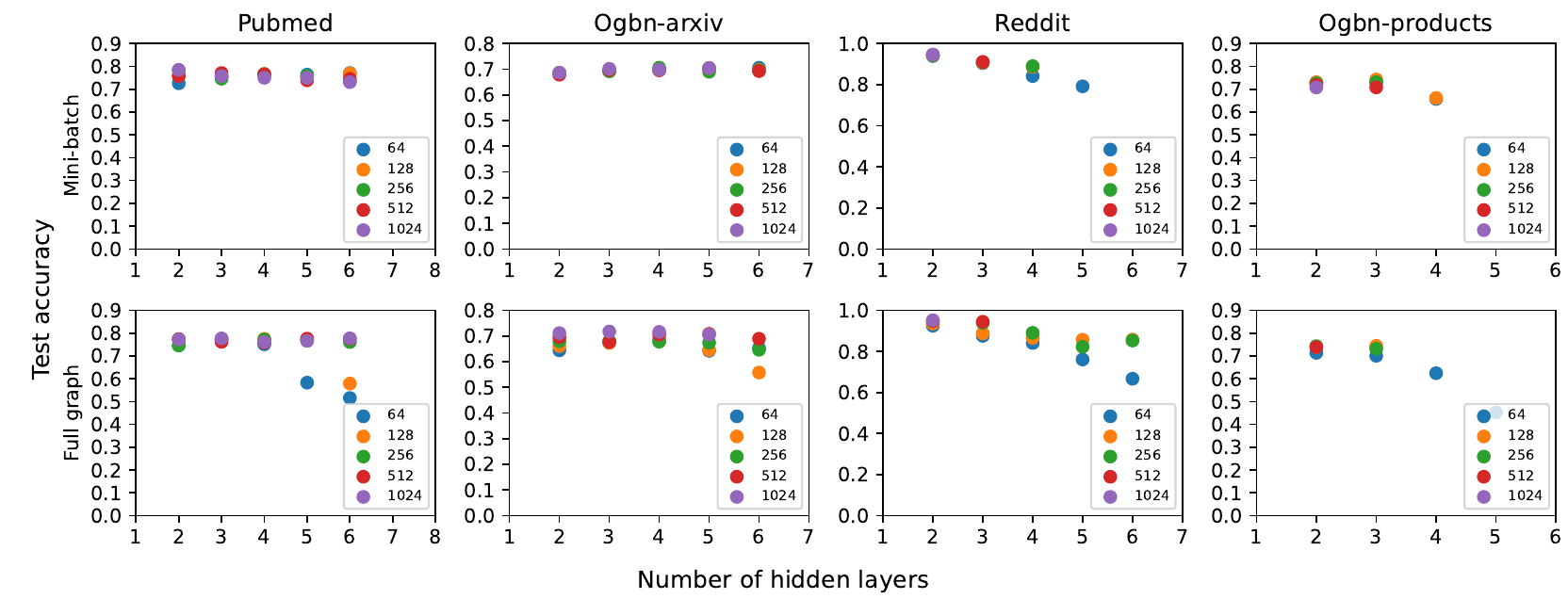}
    \vskip -0.1in
    \caption{Ablation study with the varying number of layers and hidden size for GAT model. Legend represents the size of each hidden layer. The X-axis shows the number of layers. The Y-axis shows the test accuracy. The top row is for mini-batch training and the bottom row is for full graph training. We observe that mini-batch training is robust to architectural changes, whereas full-graph training has a lot of variation.}
    \label{fig:GAT_robustness}
\end{figure*}


\emph{How sensitive are the two training methods to changes in the hyperparameters?}
We evaluate the ease of finding a good hyperparameter setting using the two training approaches by measuring how accuracy varies as we vary the model architecture hyperparameters, i.e., the number of hidden layers and the size of the hidden vertex dimensions.

We trained each model for a maximum of 1000 epochs, stopping earlier if the validation accuracy doesn't increase for 50 epochs.

Figure~\ref{fig:all_robustness} illustrates the best test accuracy when training the GraphSAGE model, the top row shows mini-batch training experiments and the bottom row shows synchronous full-graph training. Notably, mini-batch training demonstrates remarkable consistency across varying model architectures.  In contrast, full-graph training exhibits a wider range of variations in test accuracy when the architecture changes.
We observed similar results for GCN and GAT models, which are reported in Figure \ref{fig:GCN_robustness} and Figure \ref{fig:GAT_robustness}.

Figure~\ref{fig:all_robustness} and the other results for GCN and GAT, 
clearly show the effect of over-smoothing, which appears in GNNs at increased depths~\cite{rusch_a}. 
Mini-batch training on GraphSAGE does not show smoothing with up to 8 layers in Figure~\ref{fig:all_robustness}.
In a set of separate experiments, we found that over-smoothing does occur with mini-batch training, but only after we increase the number of layers to $16$ or higher. 
Our experiments for GCN and GAT show that over-smoothing occurs at lower depths for mini-batch training but the resulting accuracy drop is still much less than that with full-graph training (see the Appendix).

\spara{Takeaway.}
These results highlight the robustness of mini-batch training to alterations in model architecture.
Full-graph training appears to be more sensitive to changes in architecture, so it requires a more careful and potentially expensive hyperparameter search. Whereas mini-batch training is consistently less sensitive to changes of those parameters than full-graph training, across all the GNN models we consider. In addition, mini-batch training is much less prone to the over-smoothing problem \cite{rusch_a} that affects deep GNN models with many layers.

\begingroup
\renewcommand{\arraystretch}{0.8}
\begin{table}[t]
\caption{Time to accuracy and epoch time for GraphSAGE}
\label{tab:time_to_acc_graphsage_samplers}
\tabcolsep=0.135cm 
    \begin{center}
    \begin{small}
    \begin{tabular}{ |m{8em} |m{9em}  |>{\centering\arraybackslash}m{4.0em} |>{\centering\arraybackslash}m{5.0em} | >{\centering\arraybackslash}m{4.0em} |>{\centering\arraybackslash}m{6.0em}|} 
\toprule
 & Sampler & pubmed  & ogbn-arxiv & reddit & ogbn-products \\
\midrule

\multirow{3}{8em}{\text{Epoch time (s)}} 
& Cluster GCN      & 0.0118  & 0.0254  & 0.2798  & 0.1212  \\
& NS & 0.0120  & 0.1592  & 2.6728  & 0.5552  \\
& Saint Sampler    & 0.0123  & 0.1029  & 0.4689  & 0.0908  \\
\midrule

\multirow{3}{8em}{\text{Time to accuracy (s)}} 
& Cluster GCN      & 2.5418  & 9.0578   & 97.6485  & 111.46  \\
& NS & 1.1168  & 48.3875  & 192.440  & 122.71  \\
& Saint Sampler    & 1.0939  & 61.2358  & 151.910  & 42.9602 \\
      \bottomrule
\end{tabular}
    \end{small}
    \end{center}
\end{table}
\endgroup

\begingroup
\renewcommand{\arraystretch}{0.8}
\begin{table}[t]
\caption{Time to accuracy and epoch time for GAT}
\label{tab:time_to_acc_gat_samplers}
\tabcolsep=0.135cm 
    \begin{center}
    \begin{small}
    \begin{tabular}{ |m{8em} |m{9em}  |>{\centering\arraybackslash}m{4.0em} |>{\centering\arraybackslash}m{5.0em} | >{\centering\arraybackslash}m{4.0em} |>{\centering\arraybackslash}m{6.0em}|} 
\toprule
 & Sampler & pubmed  & ogbn-arxiv & reddit & ogbn-products \\
\midrule

\multirow{3}{8em}{\text{Epoch time (s)}} 
& Cluster GCN      & 0.0353  & 0.0621  & 1.3446  & 0.4893  \\
& NS & 0.0222  & 0.2927  & 18.5336 & 1.2580  \\
& Saint Sampler    & 0.0224  & 0.1176  & 0.3653  & 0.0524  \\
\midrule

\multirow{3}{8em}{\text{Time to accuracy (s)}} 
& Cluster GCN      & 2.3293  & 15.3941  & 599.686  & 1109.32  \\
& NS & 3.8463  & 46.2388  & 1853.36  & 159.769  \\
& Saint Sampler    & 2.4866  & 47.759   & 115.419  & 20.4534  \\
      \bottomrule
\end{tabular}
    \end{small}
    \end{center}
\end{table}
\endgroup

\begingroup
\renewcommand{\arraystretch}{0.8}
\begin{table}[t]
\caption{Time to accuracy and epoch time for GCN}
\label{tab:time_to_acc_gcn_samplers}
\tabcolsep=0.135cm 
    \begin{center}
    \begin{small}
    \begin{tabular}{ |m{8em} |m{9em}  |>{\centering\arraybackslash}m{4.0em} |>{\centering\arraybackslash}m{5.0em} | >{\centering\arraybackslash}m{4.0em} |>{\centering\arraybackslash}m{6.0em}|} 
\toprule
 & Sampler & pubmed  & ogbn-arxiv & reddit & ogbn-products \\
\midrule

\multirow{3}{8em}{\text{Epoch time (s)}} 
& Cluster GCN      & 0.0112  & 0.0259  & 0.2333  & 0.1030  \\
& NS & 0.0133  & 0.1814  & 4.4802  & 0.6186  \\
& Saint Sampler    & 0.0126  & 0.1016  & 0.4506  & 0.0905  \\
\midrule

\multirow{3}{8em}{\text{Time to accuracy (s)}} 
& Cluster GCN      & 2.5933  & 9.1563   & 39.6631  & 47.0832  \\
& NS & 1.8819  & 27.2038  & 488.345  & 64.3342  \\
& Saint Sampler    & 2.5587  & 51.6232  & 159.962  & 29.0637  \\
      \bottomrule
\end{tabular}
    \end{small}
    \end{center}
\end{table}
\endgroup

\end{document}